\documentclass[letterpaper, 10 pt, conference]{bib/ieeeconf} 
\IEEEoverridecommandlockouts       
\usepackage[normalem]{ulem}	                        
\usepackage[table,usenames,dvipsnames]{xcolor}      
\usepackage{extarrows}                              
\let\labelindent\relax
\usepackage{enumitem}                               

\usepackage{amsmath,amsthm,amssymb,amsfonts,dsfont} 
\usepackage{cite}
\makeatletter
\newif\if@restonecol
\newif\foralgo@restonecol
\makeatother

\usepackage[linesnumbered,ruled,vlined]{algorithm2e}
\usepackage{algpseudocode}
\usepackage{graphicx}
\usepackage{subfigure}
\usepackage{tabularx}
\usepackage{multirow,multicol, array}
\usepackage[font={small}]{caption} 
\usepackage{todonotes}
\let\appendices\relax
\usepackage[titletoc,title]{appendix}
\usepackage[breaklinks=true, colorlinks, bookmarks=true, citecolor=Black, urlcolor=Violet,linkcolor=Black]{hyperref}
\usepackage{cleveref}

\usepackage{setspace}
\newcommand{\tabincell}[2]{\begin{tabular}{@{}#1@{}}#2\end{tabular}}  

\theoremstyle{definition}

\newtheorem{proposition}{Proposition}

\newtheorem*{assumption*}{Assumption}
\newtheorem{remark}{Remark}
\newtheorem*{remark*}{Remark}
\newtheorem*{problem*}{Problem}

\newcommand{\x}{\mathbf{x}}
\newcommand{\y}{\mathbf{y}}

\newcommand{\argmax}{\operatorname{argmax}}
\newcommand{\argmin}{\operatorname{argmin}}

\newcommand{\xucb}{\x^{\text{ucb}}}
\newcommand{\ymax}{\y^{\max}}
\newcommand{\eymax}{{\tilde \y}^{\max}}

\newcommand{\batchx}{X}
\newcommand{\batchy}{Y}
\SetKwInOut{Parameter}{Initialization}

\newcommand{\algcomments}[1]{\textit{\texttt{\# #1}}}

\title{\LARGE \bf
Gaussian Max-Value Entropy Search for Multi-Agent Bayesian Optimization
}
\author{Haitong Ma, Tianpeng Zhang, Yixuan Wu, Flavio P. Calmon, Na Li
\thanks{The authors are with School of Engineering and
Applied Sciences, Harvard University. Email: \texttt{\{haitongma, tzhang, florencewu,fcalmon\}@g.harvard.edu, nali@seas.harvard.edu}. The work is supported under ONR YIP: N00014-19-1-2217, NSF CNS: 2003111, and NSF AI institute: 2112085.}
}

\begin{document}
\maketitle
\begin{abstract}

We study the multi-agent Bayesian optimization (BO) problem, where multiple agents maximize a black-box function via iterative queries. 
We focus on Entropy Search (ES), a sample-efficient BO algorithm that selects queries to maximize the mutual information about the maximum of the black-box function. 
One of the main challenges of ES is that calculating the mutual information requires computationally-costly approximation techniques. For multi-agent BO problems, the computational cost of ES is exponential in the number of agents. To address this challenge, we propose the \emph{Gaussian Max-value Entropy Search}, a multi-agent BO algorithm with favorable sample and computational efficiency. The key to our idea is to use a normal distribution to approximate the function maximum and calculate its mutual information accordingly. The resulting approximation allows queries to be cast as the solution of a closed-form optimization problem which, in turn, can be solved  via a modified gradient ascent algorithm and scaled to a large number of agents.
We demonstrate the effectiveness of Gaussian max-value Entropy Search through numerical experiments on standard test functions and real-robot experiments on the source seeking problem. Results show that the proposed algorithm outperforms the multi-agent BO baselines in the numerical experiments and can stably seek the source with a limited number of noisy observations on real robots.
\end{abstract}
\section{Introduction}
Bayesian optimization (BO) is a sample-efficient method for maximizing expensive-to-evaluate black-box functions, which frequently arise in robotic applications. By iteratively evaluating the black-box function at the \emph{query points}, BO first builds a probabilistic model about the function and then infers the location of the maximum accordingly. Scenarios where BO is applicable include tuning the parameters of controllers \cite{berkenkamp2016safe} and motion planners \cite{zhao2022model,ma2022joint}, seeking the location of source signal \cite{zhang2021source,zhang2022distributed}, designing the morphology structure of robots \cite{rosser2020sim2real}, and so on.

Query point selection is a fundamental challenge in BO. The key is to carefully choose the query points to learn about the objective function (exploration) while leveraging existing knowledge to maximize it (exploitation). One approach of BO is to select query points in each iterative step to maximize the so-called \emph{acquisition function}. The acquisition function regulates the tension between exploring versus exploiting and is updated at each iteration based on all the queries and observations collected so far. 
One of the most notable acquisition functions is the Gaussian process upper confidence bound (GP-UCB) \cite{auer2002using,srinivas2009gaussian} extended from the multi-armed bandit problem. GP-UCB uses a weighted sum of the posterior mean (exploitation) and the posterior variance (exploration) of the Gaussian process model to handle the trade-off.
Other popular acquisition functions include the probability of improvement (PI) \cite{kushner1964new}, expected improvement (EI) \cite{movckus1975bayesian}, knowledge gradient (KG) \cite{wu2017bayesian}, Entropy Search (ES) \cite{hennig_entropy_2012}, Thompson sampling (TS) \cite{thompson1933likelihood}, etc. 

While most BO methods use a single agent for querying, it is more desirable to have multiple agents querying the black-box function simultaneously in many applications. With multiple agents, the querying is parallelized so that more information can be obtained per iteration, and thus the objective function can be learned and maximized faster. 
Existing multi-agent BO studies have proposed two approaches to selecting the batch of query points for the agents at each iterative step: (1) sequential query calculation and (2) batch query calculation. Sequential query calculation computes the query of each agent within the batch one by one. A single-agent BO algorithm usually determines the first agent's query. Then, other agents' queries are added sequentially to provide more exploration \cite{garcia2019fully, contal2013parallel,shahriari2016unbounded} or exploitation \cite{desautels2014parallelizing,snoek2012practical}. In contrast, batch query calculation computes the batch of queries simultaneously, for example, using Thompson sampling or Entropy Search \cite{kandasamy2018parallelised,shah2015parallel,wu2016parallel}. Apart from the distinctions in the computational procedure, these two approaches also address the collaboration among the agents differently. The collaboration can be viewed as balancing the exploration-exploitation trade-off within the same batch of query points. While the sequential query calculation usually assigns explicit roles of exploration/exploitation to each agent, the batch query calculation handles the collaboration implicitly through its probabilistic model.

Among these multi-agent methods, the Entropy Search (ES) has gained increasing attention because of its promising low-regret performance \cite{hennig_entropy_2012,hennig2011optimal,wang_max-value_2017,hernandez-lobato_predictive_2014,tu_joint_2022}. ES maximizes the black-box function by maximizing the \emph{mutual information} of the estimated function maximum, which is shown to be more sample-efficient than directly querying the function at the estimated function maximum \cite{hennig_entropy_2012}. Moreover, ES is especially suitable for the multi-agent BO setting since the collaboration among the agents can be encouraged by maximizing the total mutual information about the function maximum in the agents' queries. In this way, ES can automatically adapt the multi-agent collaboration under different objective functions to obtain the most informative queries, which is more flexible and efficient than the fixed role assignment scheme used in the sequential query calculation methods. 

One of the main challenges of ES, especially for multi-agent BO problems, is computational efficiency. 
Computing the entropy of the function maximum is generally intractable and requires sophisticated approximation techniques. 
Existing methods, including Monte-Carlo sampling \cite{hennig2011optimal}, expectation propagation \cite{hennig_entropy_2012,shah2015parallel,tu_joint_2022}, random feature sampling \cite{shah2015parallel,hernandez-lobato_predictive_2014,wang_max-value_2017}, and Gambel sampling \cite{wang_max-value_2017}, have exponential computational cost in the number of agents \cite{shah2015parallel,tu_joint_2022}. 
Although \cite{shah2015parallel} proposes a gradient-based multi-agent ES algorithm to reduce the exponential cost to polynomial cost, the computation is still heavy since the gradient calculation involves a large number of matrix inversions. The required number of inversions is proportional to the dataset size.

\textbf{Our Contributions.} 
We proposed the Gaussian Max-value Entropy Search (GMES), a computationally efficient multi-agent Entropy Search algorithm with a novel entropy approximation scheme and practical implementations for the multi-agent setting.   
Specifically, we use the normal distribution to approximate the distribution of the function maximum and calculate its entropy. 
We use the mutual information for this approximate distribution as our acquisition function, which \emph{has a closed-form expression}.   
Unlike existing multi-agent Entropy Search algorithms, we do not need costly sampling when calculating the acquisition function.
We further use gradient ascent to compute the query points and add log-barrier safety constraints to make the proposed algorithm scalable to a large number of agents and applicable to real-world applications. 
Together, our algorithm has favorable computational efficiency compared to existing methods. 
    
We then test the algorithms on both numerical and real-robot experiments. Experiment results show that the proposed algorithm outperforms the baseline multi-agent BO algorithms with different numbers of agents in numerical tests. The real-robot experiments demonstrate that our algorithm can successfully seek a light source with a small number of queries. These source seeking experiments also showcase the substantial advantage of using multiple agents over a single agent. Compared to single-agent seeking, four agents improved the source-seeking time by 59.9\% and the source-seeking iterations by 67.6\% on average.

\section{Preliminaries}
\subsection{Problem Formulation}
We consider a multi-agent BO problem that consists of a team of $m$ agents, a compact domain $\mathcal{X}\subset\mathbb{R}^d$, and an \textit{unknown} function 
$$ f: \mathcal{X} \rightarrow \mathbb{R}.$$
We assume $f$ is continuous on $\mathcal{X}$, so its maximum exists in this domain. The goal is to maximize $f$ only through the queries of function values. We assume each agent can query $f$ at any point $\x\in \mathcal{X}$ and observe a noisy function value $$\y = f(\x)+\epsilon,~\epsilon\sim \mathcal{N}(0,\sigma_0^2),$$ where $\sigma_0^2$ is the variance of observation noise $\epsilon$. 

The agents query $f$ through a sequence of iterations $t=1,2,..., T$. For all agents $i\in \{1,2,...,m\}$, we use lower-case letters $\x_t^i$ for the query point of agent $i$ at time $t$ and $\y_t^i$ for agent $i$'s observation at time $t$. We use capital letters $\batchx_t \equiv \{\x_t^1,\x_t^2,\dots,\x_t^m\}$ and $\batchy_t \equiv \{\y_t^1,\y_t^2,\dots,\y_t^m\}$ to denote the collections of agents' query points and evaluations at time $t$. We denote $\mathcal{X}^m=\underbrace{\mathcal{X}\times\mathcal{X}\times\dots\mathcal{X}}_{m}\in\mathbb{R}^{md}$ as the domain of batch queries. 
We define $\mathbb{X}_t\equiv\{\batchx_1,\batchx_2,\dots,\batchx_t\}$ and $\mathbb{Y}_t\equiv\{\batchy_1,\batchy_2,\dots,\batchy_t\}$ as all the queries and observations up to time $t$. 
Let $D_t=\mathbb{X}_{t-1}\cup \mathbb{Y}_{t-1}$ be the observed data before time $t$.

\subsection{Gaussian Process}
We briefly introduce the Gaussian process (GP), our probabilistic model of the objective function. A GP model is built upon the observed data $D_t$ and a positive-definite kernel function $k(\x,\x')$ that models our prior belief about the coupling between $f(\x)$ and $f(\x')$\cite{williams2006gaussian}. Given $D_t$ and $k$, the GP model for function $f$ can be fully described by its mean value function $\mu_t(\x)$ and covariance function $\Sigma_t(\x,\x')$, which are calculated by
\begin{equation}
    \begin{aligned}
\mu_t(\x) & =\boldsymbol{k}_t(\x)^\top\left(\boldsymbol{K}_t+\sigma_0^2 \boldsymbol{I}\right)^{-1} \boldsymbol{y}_t, \\
\Sigma_t\left(\x, \x^{\prime}\right) & =k\left(\x, \x^{\prime}\right)-\boldsymbol{k}_t(\x)^\top\left(\boldsymbol{K}_t+\sigma_0^2 \boldsymbol{I}\right)^{-1} \boldsymbol{k}_t\left(\x^{\prime}\right)
\end{aligned}\label{eq:gp}
\end{equation}
\noindent where $\boldsymbol{k}_t(\x)=\left[k\left(\x', \x\right) \right]^\top_{\x'\in\mathbb{X}_{t-1}}$ and $\boldsymbol{y}_t=\left[\y\right]_{\y\in \mathbb{Y}_{t-1}}^\top$ are $m\cdot(t-1)$ dimensional vectors, and $\boldsymbol{K}_t = \left[k\left(\x, \x^{\prime}\right)\right]_{\x, \x^{\prime} \in \mathbb{X}_{t-1}}$ is a $m\cdot(t-1)\times m\cdot(t-1)$ positive definite matrix. $\boldsymbol{I}$ is the identity matrix with the same shape as $\boldsymbol{K}_t$. We denote the resulting GP model as $GP(\mu_t,\Sigma_t \mid D_t)$. We also define $\sigma_t^2(\x) = \Sigma_t(\x,\x)$ as the variance function, and let $\sigma_t(\x) = \sqrt{\sigma_t^2(\x)}$. The value $\mu_t(\x)$ can be understood as the GP model's predicted value of $f(\x)$ given the observed data $D_t$ and $\sigma_t^2(\x)$ is the uncertainty in that prediction in the form of variance. A more comprehensive discussion of topics related to GP can be found in \cite{williams2006gaussian}.

\subsection{Posterior Max-value and Entropy Search}
From the function space viewpoint, GPs can be regarded as distributions over functions \cite{williams2006gaussian}. Denote the random function sampled from posterior GP distribution conditioned on the observed data $D_t$ by $\hat{f}_t\mid D_t\sim GP(\mu_t,\Sigma_t\mid D_t)$. The corresponding maximal value of $\hat{f}_t$ is thus a random variable, which we denote as
\begin{equation}
    \ymax\mid D_t\sim\max_{\x\in\mathcal{X}}\hat{f}_t (\x) \mid D_t
\end{equation}
We name $ \ymax\mid D_t$ as the \emph{posterior max-value}. 
The key of Entropy Search (ES) is to select query points that maximally reduce the uncertainty in $\ymax$, and specifically, the uncertainty is quantified by the differential entropy of $\ymax$, defined by
\begin{equation}
    H(\ymax\mid D_t) =- \int p(\ymax\mid D_t)\log(p(\ymax\mid D_t))\ .\label{eq:ymaxent}
\end{equation}

The uncertainty about $\ymax$ decreases after gaining information from the new queries and observations $\{\batchx_t,\batchy_t\}$. The reduction in uncertainty is called \emph{mutual information}. Mutual information
is formally defined by the entropy reduction after more information is given,
\begin{align}
       & I(\ymax;\{\batchx_t,\batchy_t\}\mid D_t)\label{eq:defmi} \\
    =& H(\ymax\mid D_t) - H(\ymax\mid D_t\cup\{\batchx_t,\batchy_t\})\label{eq:mesacq1} \\
    =& H(\batchx_t,\batchy_t \mid D_t) - H(\batchx_t,\batchy_t\mid D_t,\ymax) \label{eq:mesacq2}
\end{align}
where $I(\ymax;\{\batchx_t,\batchy_t\}\mid D_t)$ denotes the mutual information of $\ymax$ after new data $\{\batchx_t,\batchy_t\}$ is added.
Equation \eqref{eq:mesacq1} follows from the definition of mutual information and
Equation \eqref{eq:mesacq2} is from the symmetric property of mutual information \cite{cover1999elements}. In summary, ES uses the mutual information above as the acquisition function to quantify the uncertainty reduction about $\ymax$. The agents query $\batchx_t$ that maximizes \eqref{eq:defmi} to get the most information about $\ymax$.

The key challenge in ES is \emph{how to calculate the mutual information in \eqref{eq:defmi}} since it usually does not have a closed form. Equations \eqref{eq:mesacq1} and \eqref{eq:mesacq2} represent two different approaches, and both have difficulties. The difficulty in \eqref{eq:mesacq1} is that the differential entropy $H(\ymax\mid D_t\cup\{\batchx_t,\batchy_t\})$ is conditioned on $\batchy_t$ which is not known before querying the function at $\batchx_t$\cite{hennig2011optimal,hennig_entropy_2012}. One solution proposed by \cite{seeger2004gaussian} is to sample a batch of realizations $\{\hat{\batchy}_{1,t},\hat{\batchy}_{2,t},...\}$ from the current posterior distribution $\hat f(\batchx_t)\mid D_t$ and compute $H(\ymax\mid D_t\cup\{\batchx_t,\batchy_t\})$ by averaging over the samples $\{H(\ymax \mid D_t \cup \{\batchx_t,\batchy_t=\hat{\batchy}_{i,t}\}): i=1,2,...\}$. 
However, the corresponding computational cost is high since for each $\hat{\batchy}_{i,t}$ we need to calculate $H(\ymax \mid D_t \cup \{\batchx_t,\batchy_t=\hat{\batchy}_{i,t}\})$ once, and the calculation again requires costly sampling. Therefore, the total computation for \eqref{eq:mesacq1} is substantial.
 
To avoid conditioning on $\batchy_t$, recent studies have considered computing the mutual information with equation \eqref{eq:mesacq2} \cite{hernandez-lobato_predictive_2014,shah2015parallel,wang_max-value_2017,tu_joint_2022}. However, this approach still requires sophisticated computation because the term $H(\batchx_t,\batchy_t\mid D_t,\ymax)$ is non-trivial due to the conditioning on $\ymax$. 
 
The multi-agent nature of our problem brings further computation challenges to ES. Computing the mutual information through either Eq. \eqref{eq:mesacq1} or Eq. \eqref{eq:mesacq2} typically requires averaging over a sufficient number of samples from the $m$-dimensional posterior distribution $\batchy_t\mid D_t$. Therefore, the number of samples needed for a good approximation of the mutual information is exponential in the number of agents $m$. Furthermore, optimizing the non-trivial entropy over $X_t$ on the $m\times d-$dimensional space introduces even more difficulties\cite{tu_joint_2022}. Brute-force methods, such as grid or random search, also induce  exponential optimization cost (in $m$).

In this paper, we follow the approach of \eqref{eq:mesacq1} to calculate the acquisition function. We aim to reduce the computation cost and leverage the information-theoretic collaboration scheme to develop an efficient multi-agent Entropy Search algorithm that is practical for real robotic applications. We use the normal distribution to approximate the distribution of $\ymax$ and use the mutual information associated with the approximated distribution as our acquisition function. We show that our acquisition function has an explicit expression; its computation is thus free from the expensive sampling over the $m$-dimensional distribution $Y_t\mid D_t$ in previous approaches like \cite{seeger2004gaussian}. 
With the closed-form acquisition function, we develop a centralized multi-agent Entropy Search algorithm where multiple agents collaborate on a BO task. 

\section{Gaussian Max-Value Entropy Search}

\subsection{Gaussian Approximation of Posterior Max-Value}

The differential entropy of $\ymax$ in Eq. \eqref{eq:ymaxent} usually do not have a closed form. To overcome this difficulty, we use the distribution of a Gaussian random variable $\eymax$ to approximate the posterior distribution $\ymax\mid D_t$ so that the resulting mutual information associated with $\eymax$ has a closed-form expression. The idea is reasonable since the maximal value of real function $f$ should be a deterministic value. If we further consider the noise $\epsilon$, then the real observed max-value distribution should be Gaussian.

We formally define $\eymax$ as follows. Denote the location with the maximal upper confidence bound(UCB) value at time $t$ as
$$\xucb_t\equiv\argmax_{\x \in \mathcal{X}} (\mu_t(\x) + \beta_t\sigma_t(\x))$$
where $\beta_t$ is a hyper-parameter indicating the confidence interval length. The UCB, $\mu_t(\x) + \beta_t\sigma_t(\x)$, follows the definition in GP-UCB\cite{srinivas2009gaussian}. Define the \textit{estimated} posterior max-value $\eymax\mid D_t$ as
\begin{equation}
    \begin{aligned}
        \eymax\mid D_t &\sim\mathcal{N}(\mu_{t}(\xucb_t), \sigma_{t}^2(\xucb_t))\\
        \eymax\mid D_t \cup\{\batchx_t,\batchy_t\} &\sim\mathcal{N}(\mu_{t+1}(\xucb_t), \sigma_{t+1}^2(\xucb_t))
    \end{aligned}
\end{equation}
Here, $\mu_{t}(\xucb_t)$, $\sigma_{t}^2(\xucb_t)$ and $\mu_{t+1}(\xucb_t)$ and $\sigma_{t+1}^2(\xucb_t)$ are the $t$ and $t+1$ step posterior mean and variance functions  at $\xucb$. $\mu_{t}(\xucb_t)$, $\sigma_{t}^2(\xucb_t)$ are computed before we get the observation $\batchy_t$ at $\batchx_t$, and $\mu_{t+1}(\xucb_t)$,  $\sigma_{t+1}^2(\xucb_t)$ are computed after we get the observation $\batchy_t$ at $\batchx_t$. This way, we can compare the entropy before and after observing $\batchy_t$ to calculate the mutual information.

We approximate the mutual information in \eqref{eq:defmi} with $I(\eymax;\{\batchx_t,\batchy_t\}\mid D_t)$, which is given by  
\begin{align}
    &I(\eymax;\{\batchx_t,\batchy_t\}\mid D_t)\\
    =& H(\eymax\mid D_t)-  H(\eymax\mid D_t \cup\{\batchx_t,\batchy_t\} )\label{eq:approx-acqt} \\ 
    =& \frac{1}{2}\log(2\pi\sigma_{t}^2(\xucb_t)) - \frac{1}{2}\log(2\pi\sigma_{t+1}^2(\xucb_t)) \label{eq:approx-acq}
\end{align}
where we use the differential entropy of normal distribution to convert Eq. \eqref{eq:approx-acqt} to Eq. \eqref{eq:approx-acq}.

We propose to use Eq. \eqref{eq:approx-acq} as the surrogate of the mutual information in \eqref{eq:defmi}. Since $\sigma_t^2$ is independent of $\{\batchx_t,\batchy_t\}$, the first term in \eqref{eq:approx-acq} is independent of $\batchx_t$ and can be omitted, thus to maximize \eqref{eq:approx-acq} in $X_t$ is to minimize $ \frac{1}{2}\log(2\pi\sigma_{t+1}^2(\xucb_t))$, or equivalently, $\sigma_{t+1}^2(\xucb_{t})$. The explicit expression of $\sigma_{t+1}^2(\xucb_{t})$ is given by the following proposition.

\begin{proposition}[Predicted change of GP posterior variance]\label{prop:gamma}
The posterior variances at any point $\x$ at times $t$ and $t+1$ are related by
\begin{equation}
    \begin{aligned}
        &\sigma^2_{t+1}(\x) =\sigma^2_{t}(\x) -  \gamma(\batchx_t,\x)
\end{aligned}
\end{equation}

where
\begin{equation}
    \begin{aligned}
        &\gamma(\batchx_t,\x)\\
        \equiv &\Sigma_t\left(\x, \batchx_t\right) (\Sigma_t(\batchx_t, \batchx_t)+\sigma_0^2 I_m)^{-1}\Sigma_t\left(\batchx_t, \x\right)\ . \label{eq:gammaxt}
    \end{aligned}
\end{equation}
Here, $I_m$ is the $m$-dimensional identity matrix, $\Sigma_t(\x,\batchx_t)$ stands for the row vector  $$\Sigma_t(\x,\batchx_t)\equiv[\Sigma_t(\x,\x_t^1),\Sigma_t(\x,\x_t^2),\dots,\Sigma_t(\x,\x_t^m)],$$
while $\Sigma_t(\batchx_t,\x) = \Sigma_t(\batchx_t,\x)^\top$, and $\Sigma_t(\batchx_t, \batchx_t)=[\Sigma_t(\x, \x')]_{\x, \x'\in \batchx_t}$ is a $m\times m$ matrix.

\end{proposition}
The proof of the proposition can be found in Appendix A of our online report\cite{online_report}. 
Proposition \ref{prop:gamma} implies that to maximize the surrogate mutual information \eqref{eq:approx-acq} in $\batchx_t$ is to maximize $\gamma(\batchx_t,\xucb_t)$.
Note that Eq. \eqref{eq:gammaxt} does not involve the observation $\batchy_t$, which is promising since it avoids the costly sampling over the $m$-dimensional distribution $Y_t\mid D_t$ to estimate $\batchy_t$ as in previous methods \cite{hennig_entropy_2012,hernandez-lobato_predictive_2014}.

The resulting multi-agent algorithm, which uses $\gamma(\cdot,\xucb_t)$ as the acquisition function to calculate batch queries $\batchx_t$, is listed in Algorithm \ref{alg:ori}. There is a central coordinator with which all agents communicate. The central coordinator receives the observations from multiple agents, updates the GP model, calculates the queries by maximizing the mutual information according to Eq. \eqref{eq:approx-acq}, and publishes the queries back to the agents. Note that Algorithm \ref{alg:ori} can be reduced to a single-agent algorithm without further changes.
\begin{algorithm}[htbp]
    \caption{Multi-Agent Gaussian Max-value Entropy Search (GMES)}\label{alg:ori}
    
    \LinesNumbered
    
    \Parameter{
    Agent number $m$, 
    Maximal iterations $T$, 
    Observation noise $\sigma_0^2$, 
    Confidence interval width $\beta_t$, 
    A central coordinator with data set $D_0 = \emptyset$, 
    initial queries $\batchx_0$,
    Gaussian process prior model $GP(\mu_0,\Sigma_0)$: $\mu_0$ is the zero function, $\Sigma_0$ is the white kernel with noise level $\sigma_0^2$. 
    }
    
     \KwOut{Inferred maximum, $\argmax_{\x\in \mathcal{X}}\mu_T(\x)$, from $GP(\mu_T,\Sigma_T\mid D_T)$.}
    
    \For{$t = 1,2,\dots,T$}
    {
    \algcomments{Each agent gets its observation at the query point}
    
        \For{Agent $i=1,2,\dots,m$}
        {
        Observe $\y_{t-1}^i = f(\x_{t-1}^i) + \epsilon$, $\epsilon\sim\mathcal{N}(0,\sigma_0^2)$ 
        }

    \algcomments{All agents return the observation to the central coordinator}
    
    $D_t = D_{t-1}\cup \{\batchx_{t-1}, \batchy_{t-1}\}$, where $\batchy_{t-1}=\{\y_{t-1}^1,\y_{t-1}^2,\dots,\y_{t-1}^m\} $

    \algcomments{Central coordinator calculates queries}
    $\xucb_t\leftarrow\argmax_{\x\in \mathcal{X}} \mu_t(\x) + \beta_t\sigma_t(\x)$
    
    $\batchx_t=\{\x_t^1,\x_t^2,\dots,\x_t^m\}\leftarrow\argmax_{\batchx\in \mathcal{X}^m} \gamma(\batchx,\xucb_t)$

    \algcomments{Central coordinator publishes queries to agents}
    
    Publish queries $\x_t^i$ to agent $i$ for $i\in\{1,2,\dots,m\}$}
\end{algorithm}

\begin{remark}\label{remark:exploration-exploitation}
It is worth discussing how our algorithm addresses the exploration-exploitation trade-off. 
In multi-agent BO, this trade-off can be balanced on two dimensions: time and batch. In our algorithm, the batch-dimension trade-off is balanced through the design of $\gamma$ while the time-dimension trade-off is guided by $\xucb_t$.

The time-dimension trade-off is that given the observations in the past, the next query points should balance between visiting the empirically good locations (exploitation) and covering under-explored locations (exploration). This trade-off is already studied intensively in single-agent BO. One of the most notable methods is to use UCB as the acquisition function, like the one in our definition of $\xucb_t$. These algorithms typically favor locations with high values in the sum of the exploitation term $\mu_t$ and the exploration term $\sigma_t$. 

The batch-dimension trade-off arises only in the multi-agent setting. It means the query points at the same iteration should remain close to some empirically high-value locations and be sufficiently diverse. This trade-off can be seen through the tension between the exploitation term $\Sigma_t(\x,\batchx_t)$ (and its transpose) and the exploration term $(\Sigma_t(X_t,X_t)+\sigma_0^2 I_m)^{-1}$ in $\gamma(X_t,\xucb_t)$. Without loss of generality, let us only consider the magnitudes of these terms. With $\gamma(X_t,\xucb_t)$ as the acquisition function, $X_t$ should be selected so that $\Sigma_t(\batchx_t,\xucb_t)$ is large, thus the query points $X_t$ should be highly correlated with $\xucb_t$, which typically implies $\x_t^1,\x_t^2,\dots,\x_t^m$ are spatially close to $\xucb_t$. Meanwhile, $X_t$ should also be chosen such that $\Sigma_t(X_t,X_t)$ is close to the zero matrix, meaning the correlation between the points in $X_t$ themselves are small, which typically implies $\x_t^1,\x_t^2,\dots,\x_t^m$ maintain some spatial separation among themselves. 
The two components of $\gamma$ thus balance the tendency of the query points to stay close to $\xucb_t$ and to maintain spatial separation among themselves, respectively.  

\end{remark}




\subsection{Practical Implementation of Algorithm \ref{alg:ori}}
A few details of Algorithm \ref{alg:ori} need to be modified for efficient computation and safety considerations in the real world. We briefly discuss the changes below. The full version of the resulting algorithm can be found in Appendix B of our online report\cite{online_report}.

\subsubsection{Calculating Queries by Gradient Ascent}
One computational challenge of Algorithm \ref{alg:ori} is to find the $\batchx \in \mathcal{X}^m$ that maximizes $\gamma(\batchx,\xucb_t)$. Brute-force methods such as grid or random search are unsuitable for multi-agent implementations as the search space grows exponentially in the number of agents $m$. 
We mitigate the expensive computation in brute-force methods and approximate the maximization of $\gamma$ through gradient ascent. Given the analytical form of $\gamma$, its gradients can be computed efficiently using standard auto-differentiation software packages like PyTorch\cite{gardner2018gpytorch}.  

To ensure the convergence of the gradient updates, an appropriate sequence of step sizes $\{\delta_t\mid t=1,2,...,T\}$ needs to be applied to the gradient. In practice, algorithms such as Adam\cite{kingma2014adam} would suffice to decide the step sizes. 

Finally, to guarantee that the algorithm returns a $X_t$ that is inside the compact domain $\mathcal{X}^m$, we project the query points back to $\mathcal{X}^m$ after every gradient update. The projection operator is defined by 
\begin{equation}
    \Gamma_{\mathcal{X}^m}(\batchx) = \underset{\bar{\batchx}\in\mathcal{X}^m}{\argmin}\|\bar{\batchx}-\batchx\|_2. \label{eq:grad_ascent}
\end{equation}

And line 9 in Algorithm \ref{alg:ori} can be replaced with multiple iterations of the gradient ascent update described below. We use $N$ to represent the number of gradient ascent iterations.
$$\batchx_t\leftarrow \Gamma_{\mathcal{X}^m}\big(\batchx_t + \delta_t\cdot\nabla_{\batchx_t}\gamma(\batchx_t, \xucb_t)\big)$$ 
 
\subsubsection{Ensuring Safety with Log-Barrier}\label{sec:safety-log-barrier}
Safety considerations, like collision avoidance, are common in many real-world multi-agent exploration tasks. Safety considerations require query points in the same batch $\batchx_t$ to be separated by at least the physical size of the robot. When this constraint is enforced, we subtract a log-barrier term $p(\batchx_t)$ from the acquisition function in \eqref{eq:gammaxt}. The log-barrier $p(\batchx_t)$ is defined by
\begin{equation}
    p(\batchx_t) = \sum_{i,j:1\leq i<j\leq m} [-\frac{1}{L}\log(\|\x_t^i-\x_t^j\| - r_{\text{div}})]^+  \label{eq:gammaxtp}
\end{equation}
where $[\cdot]^+$ means projection to the positive half-space $[0,\infty)$. 
The parameter $r_{\text{div}}$ specifies the minimal separation between the robots. We initialize $X_t$ to ensure $\|\x_t^i-\x_t^j\| > r_{\text{div}}$ for all $i\neq j$, so that $p(X_t)$ is well-defined in all iterations of the gradient updates. 

\begin{figure*}[ht]
	        \centering
    \vspace{-5pt}
        
    \includegraphics[width=0.25\linewidth]{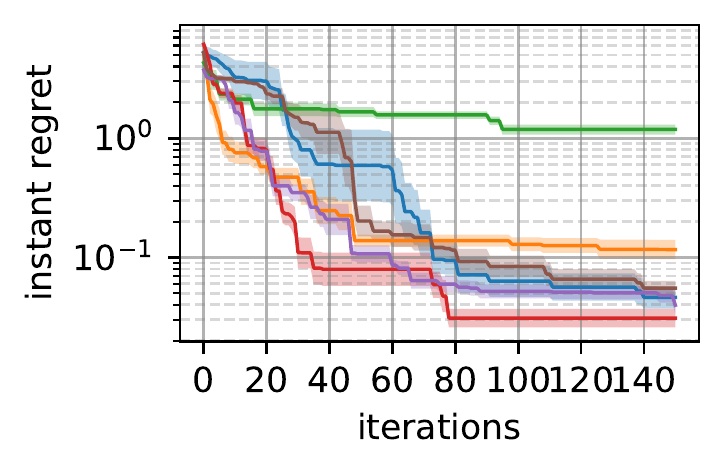}
    \includegraphics[width=0.25\linewidth]{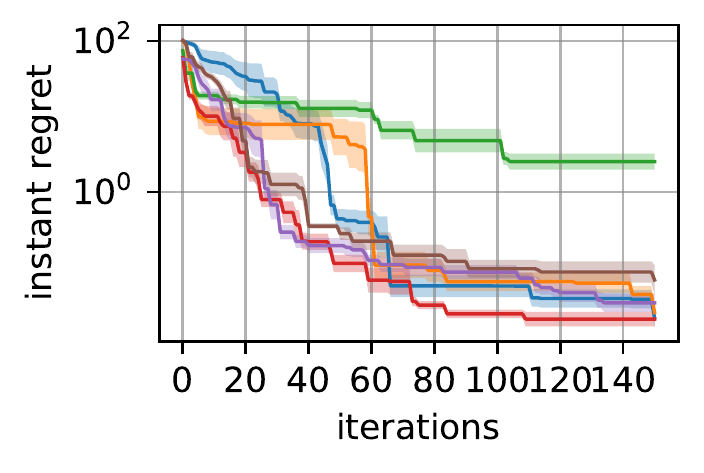}
\includegraphics[width=0.25\linewidth]{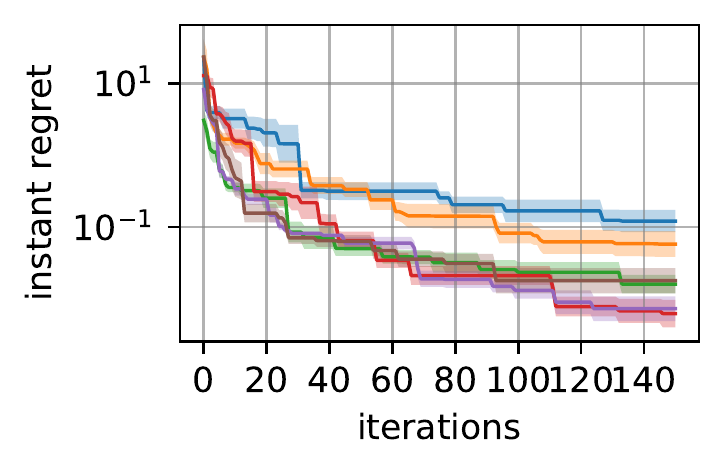}\\

    \setcounter{subfigure}{0}
    \subfigure[Ackley]{\includegraphics[width=0.25\linewidth]{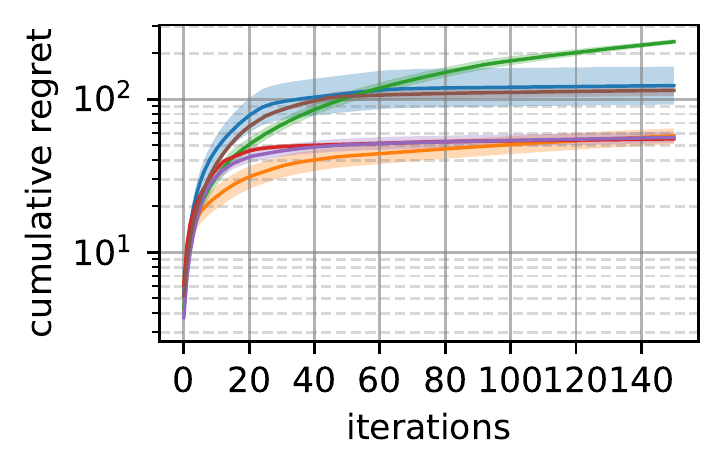}}
    \subfigure[Bird]{\includegraphics[width=0.25\linewidth]{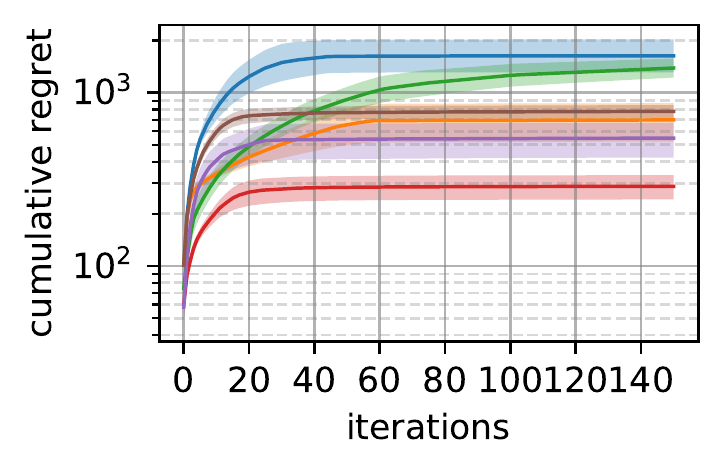}}
    \subfigure[Rosenbrock]{\includegraphics[width=0.25\linewidth]{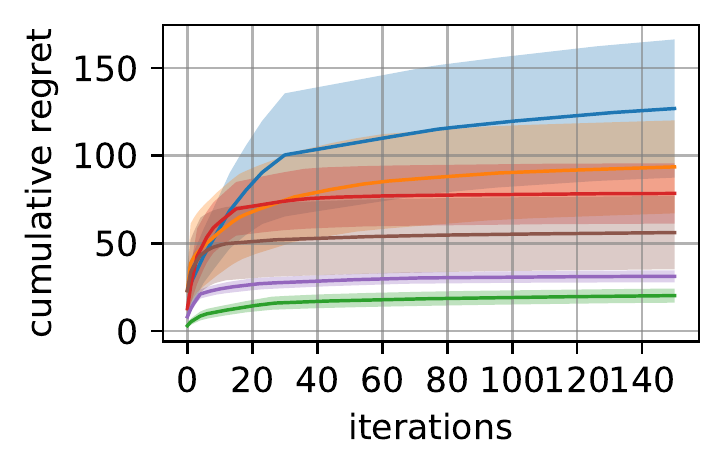}}
    \vspace{-10pt}
    \includegraphics[width=0.45\linewidth]{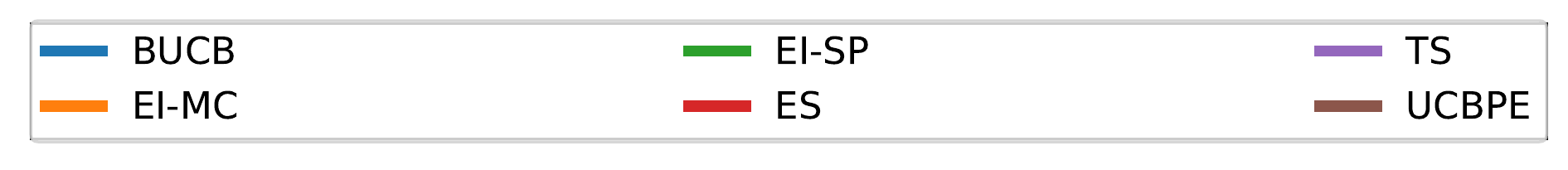}
    \caption{Instant regret (upper row) and cumulative regret (lower row). Five agents are used in all experiments. The solid lines are the average performance, and the shaded regions are the 95 percent confidence interval across five runs.} \label{fig:basic-regret}
\end{figure*}

\section{Numerical Experiments}
\label{sec:numerical}

We conduct numerical experiments with standard test functions to show the advantage of our proposed algorithm compared to recent multi-agent BO baselines. The open-source implementation of the numerical experiments can be found on \url{https://github.com/mahaitongdae/dbo}.

\begin{figure}[ht]
    \centering
    \subfigure[Ackley]{\includegraphics[width=0.22\linewidth]{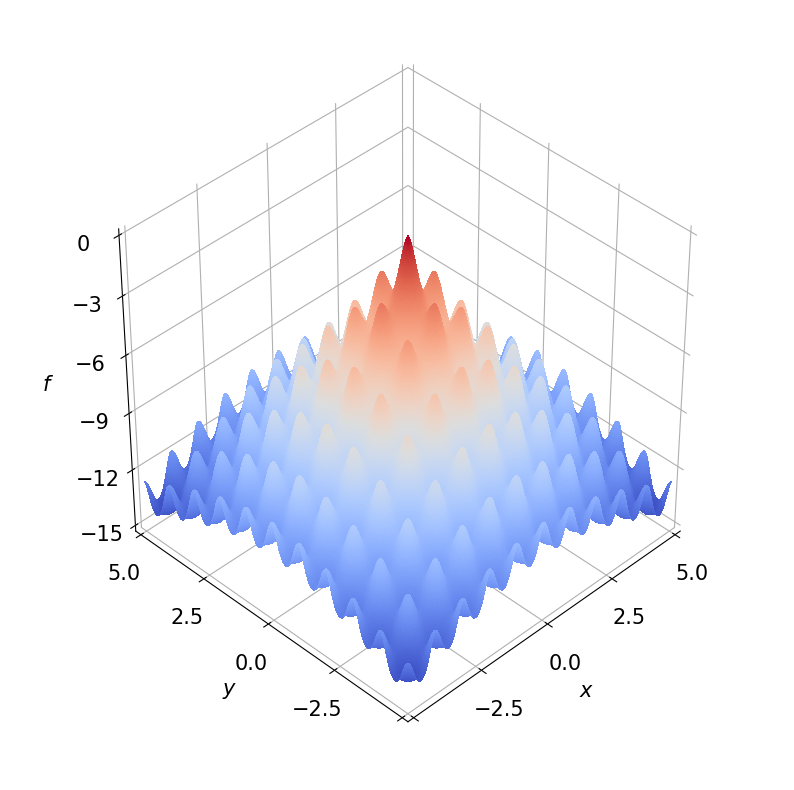}}
    \subfigure[Bird]{\includegraphics[width=0.22\linewidth]{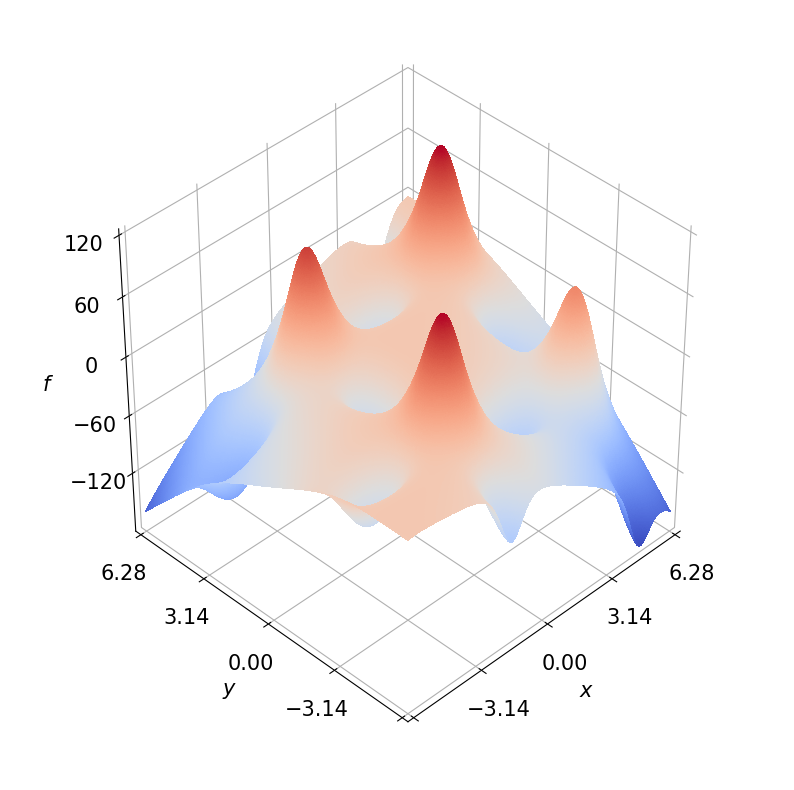}}
    \subfigure[Rosenbrock]{\includegraphics[width=0.22\linewidth]{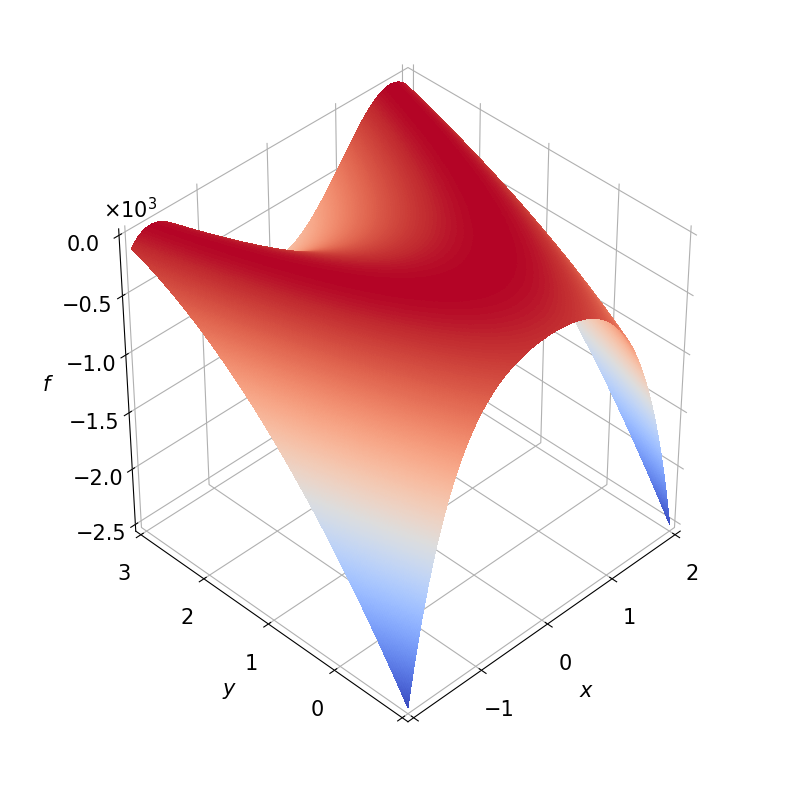}}
   \caption{Test Functions in the Numerical Experiments}
    \label{fig:objective}
\end{figure}

\subsection{Numerical Experiment Setup}
Figure \ref{fig:objective} shows the optimization landscapes of the test functions for the numerical experiments. The Ackley function has many local maxima, but only one global maximum, and its value tends to be higher at points closer to the origin. The Bird function has two global maxima and two local maxima. The Rosenbrock function has a global maximum surrounded by many saddle points. We use the Mat\'ern kernel as the kernel function $k(\cdot,\cdot)$ in \eqref{eq:gp} \cite{williams2006gaussian}. The log-barrier term is not applied to our algorithm in the numerical experiments. The algorithm parameters are listed in Table \ref{tab:algo_para}. 
\begin{table}[ht]
\centering
\caption{Algorithm parameters for the numerical experiments} \label{tab:algo_para}
\begin{tabular}{ccc}
\hline
Notation        & Meaning                                   & Value \\ \hline
$T$ & Total iterations            & 150                      \\
$\sigma_0$      & Observation noise & 0.1                      \\
$N$ & Gradient ascent iterations      & 50   \\
$\beta_t$ & Confidence interval width      & $3-0.01\cdot t$   \\
$\nu$ & Mat\'ern kernel parameter      & 1.5   \\
$p(\cdot)$ & Log-barrier term & Not applied\\
\hline
\end{tabular}
\end{table}

The algorithms are evaluated under two performance metrics, instant regret $R_t$ and cumulative regret $\bar R_t$, defined by
\begin{equation}
    R_t   =f^* -  \max_{\tau\in\{1,2,\dots,t\}}\max_{\x_\tau\in\batchx_\tau}f(\x_\tau),\quad 
        \bar R_t = \sum_{\tau=0}^{t}R_\tau
\end{equation} 
\noindent where $f^* = \max_{\x\in\mathcal{X}}f(\x)$. Note that the instant regret $R_t$ only considers the best observation among all agents over all iterations, which is a fair metric even for those algorithms that explore aggressively (like the UCB-PE). 

\subsection{Experiment Results for multi-agent BO}

We compare our algorithm with several baseline multi-agent BO algorithms from recent literature, including GP-UCB with pure exploitation (GP-UCB-PE) \cite{contal2013parallel}, Gaussian process batch upper confidence bound (GP-BUCB) \cite{desautels2014parallelizing} EI with Monte-Carlo sampling (EI-MCMC) \cite{snoek2012practical}, EI with stochastic policies (EI-SP) \cite{garcia2019fully}, and Parallel Thompson sampling (TS) \cite{thompson1933likelihood,contal2013parallel}. 

The initial query points are randomly selected using a fixed random seed in each experiment. We run each algorithm five times and report the instant and cumulative regrets for 5-agent experiments in Figure \ref{fig:basic-regret}. 
Our proposed algorithm, labeled GMES, has the lowest instant regret on all test functions with five agents. It also has the lowest cumulative regret on the Ackley and Bird functions.  EI-SP and UCB-PE achieve better cumulative regret than our algorithm on the Rosenbrock function, which shows that our algorithm might take more iterations to reach its lowest instant regret than these two baselines in this task.
Table \ref{tab:regret} shows the instant regret at the final iteration for 10- and 30-agent experiments. Our algorithm is still consistently better than the baseline algorithms in instant regrets, being the second-best on the Ackley and Rosenbrock tasks in the 30-agent experiment and the best for all other tasks. More experimental results, including comparing different numbers of agents up to 50, could be found in Appendix D in the online report \cite{online_report}.

\begin{table*}[ht]
\caption{Mean and Variance of instant regret($\times 10 ^{-2}$) at the last iteration (150 iterations in total) for 10- and 30-agent experiments. }\label{tab:regret}
\centering
\begin{tabular}{ccccccc}
\hline
           & GMES-10 (ours) & EI-SP-10 & GP-BUCB-10 & GP-UCBPE-10 & TS-10 & EI-MCMC-10 \\ \hline
Ackley     &  \tabincell{c}{\textbf{3.383} \\ $\pm$\textbf{0.503}}     &   \tabincell{c}{52.26 \\ $\pm$12.26}       &          \tabincell{c}{3.411 \\ $\pm$0.800}  &       \tabincell{c}{4.619 \\ $\pm$1.00}      &    \tabincell{c}{5.636 \\ $\pm$2.243}   &        \tabincell{c}{19.36 \\ $\pm$4.259}    \\
Bird       & \tabincell{c}{\textbf{3.626} \\ $\pm$\textbf{1.409}}      &   \tabincell{c}{33.205 \\ $\pm$11.696}       &         \tabincell{c}{5.057 \\ $\pm$2.178}   &      \tabincell{c}{2.671 \\ $\pm$0.639}       &   \tabincell{c}{5.632 \\ $\pm$1.955}    &       \tabincell{c}{3.954 \\ $\pm$1.038}     \\
Rosenbrock &  \tabincell{c}{\textbf{1.030} \\ $\pm$ \textbf{0.313}}& \tabincell{c}{11.08 \\ $\pm$3.153}    &       \tabincell{c}{21.12 \\ $\pm$10.52}     &     \tabincell{c}{1.572 \\ $\pm$0.264}         &   \tabincell{c}{41.56 \\ $\pm$7.724}   &    \tabincell{c}{1.283 \\ $\pm$0.551}        \\ \hline
& GMES-30 (ours) & EI-SP-30 & GP-BUCB-30 & GP-UCBPE-30 & TS-30 & EI-MCMC-30 \\ \hline
Ackley     & \tabincell{c}{3.279\\  $\pm$ 2.604}      &     \tabincell{c}{3.570\\ $\pm$2.300}     &   \tabincell{c}{4.670\\ $\pm$2.805}         &     \tabincell{c}{3.411\\ $\pm$2.282}        &    \tabincell{c}{\textbf{3.218}\\ $\pm$\textbf{2.089}}   &     \tabincell{c}{12.047\\ $\pm$6.094}       \\
Bird       &   \tabincell{c}{\textbf{1.857} \\ $\pm$\textbf{1.433}}    &    \tabincell{c}{2.026 \\ $\pm$1.569}      &   \tabincell{c}{2.244 \\ $\pm$1.990}         &   \tabincell{c}{2.026 \\ $\pm$1.569}          & \tabincell{c}{2.421 \\ $\pm$2.365}      &  \tabincell{c}{2.149 \\ $\pm$1.803}          \\
Rosenbrock &     \tabincell{c}{0.859 \\ $\pm$1.121}  &     \tabincell{c}{2.370\\ $\pm$1.990}     &   \tabincell{c}{49.653\\ $\pm$24.728}         &    \tabincell{c}{1.540\\ $\pm$0.659}         &   \tabincell{c}{\textbf{0.436}\\ $\pm$\textbf{0.847}}    &     \tabincell{c}{1.447\\ $\pm$2.994}       \\\hline
\end{tabular}
\end{table*}


\subsection{Analysis of Query Distributions}
\label{sec:queries}
\begin{figure*}
    \centering
    \includegraphics[width=0.42\linewidth]{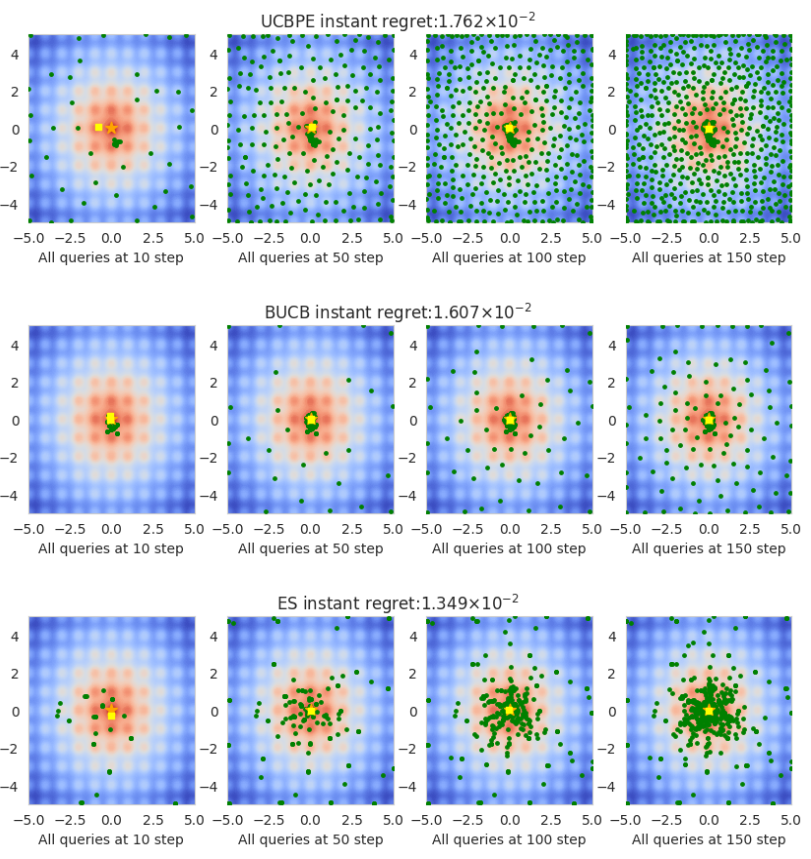}
    \includegraphics[width=0.42\linewidth]{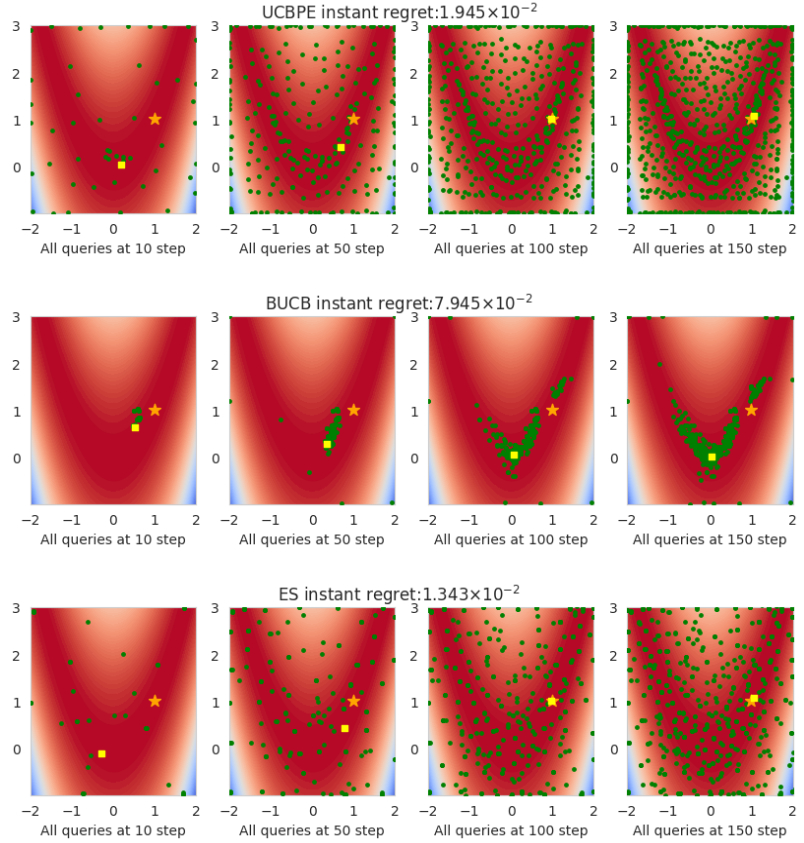}
    \includegraphics[width=0.5\linewidth]{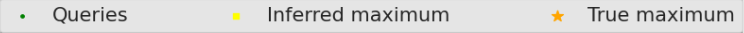}
    \caption{Query Distributions. The four columns on the left are the results of the Ackley task; the other four are the results of the Rosenbrock task. Every row indicates all the queries up to the 10, 50, 100, and 150 iterations.  In each figure: the background contour plot indicates the true objective function value; the orange star marks the global maximum; the yellow square marks the inferred maximum (the maximum location of the posterior mean function $\mu_t$) at the corresponding iteration; green dots are queries generated the algorithms; the title indicates the instant regret at the final iteration. }
    \label{fig:queries}
\end{figure*}

The following experiments demonstrate our algorithm's advantage in adapting the exploration-exploitation trade-off to different test functions. Figure \ref{fig:queries} plots the query point distributions of the proposed algorithm with gradient ascent and two baselines, BUCB and UCB-PE, that assign fixed exploration/exploitation roles to the agents. The data comes from the 10-agent experiments on Ackley and Rosenbrock tasks in the previous subsection. Our proposed algorithm (labeled as ES here) has shown \emph{the ability to adjust its exploration-exploitation balance for} different tasks. It exploits the global landscape of Ackley by clustering its query points near the origin, where the function values are generally higher. It also generates a diverse query distribution on Rosenbrock, allowing its inferred maximum (i.e., $\argmax_{\x \in \mathcal{X}} \mu_t(\x)$) to approach the true maximum as more observations are made. The final instant regrets of ES are low on both tasks. 

In contrast, the baseline algorithms do not adapt sufficiently to the two tasks. UCB-PE always has sparse query distributions, which allows it to explore the relatively flat landscape of Rosenbrock and gives it decent performance on this task; however, UCB-PE is not the best algorithm on Ackley since it does not sufficiently exploit the prominent peak at the origin.  BUCB outperforms UCB-PE on Ackley since it exploits well. However, BUCB has little diversity in its queries on Rosenbrock, making its exploration insufficient and its instant regret much higher than the other two algorithms. The inferred maximum of BUCB also gets stuck in a saddle point instead of converging to the true maximum.
This lack of diversity of BUCB's query points on Ronsenbrock could be due to BUCB's iterative greedy approach to deciding the batch of query points. 

The inflexibility of BUCB and UCB-PE is ultimately caused by their fixed exploration/exploitation role assignment regardless of the test function. In comparison, our algorithm selects query points that return the most informative queries for different tasks by maximizing the acquisition function $\gamma(\cdot,\xucb_t)$. Our algorithm can thus adapt its exploration-exploitation balance to the test functions above and outperform BUCB and UCB-PE.  

\section{Source Seeking Physical Experiments}
\label{sec:real}
We demonstrate the effectiveness of our algorithm on real robots through experiments on the multi-agent source seeking problem. The task is to let multiple robotic vehicles with sensing abilities collaborate to locate a source of interest. Examples of source seeking problems include pollutant source localization \cite{bayat2016optimal}, distributed sensor placement \cite{bachmayer2002vehicle}, target tracking \cite{morbidi2012active}, and so on. Typically, the source location is where the sensor reading is the strongest; therefore, the problem is often viewed as equivalent to driving the robots to maximize their sensor readings. Measuring the source signal is usually inexpensive, but driving the robots to desired locations could be energy- and time-consuming. Therefore, locating the source with limited samples is desirable, and BO algorithms are suitable for source seeking applications due to their sample efficiency. The full version of our experiment video can be found on \url{https://youtu.be/PK_emQ85sb0}.
\subsection{Experiment Setup}
\begin{figure}[ht]
    \centering
    \subfigure[SINGLE]{\includegraphics[height=0.25\linewidth]{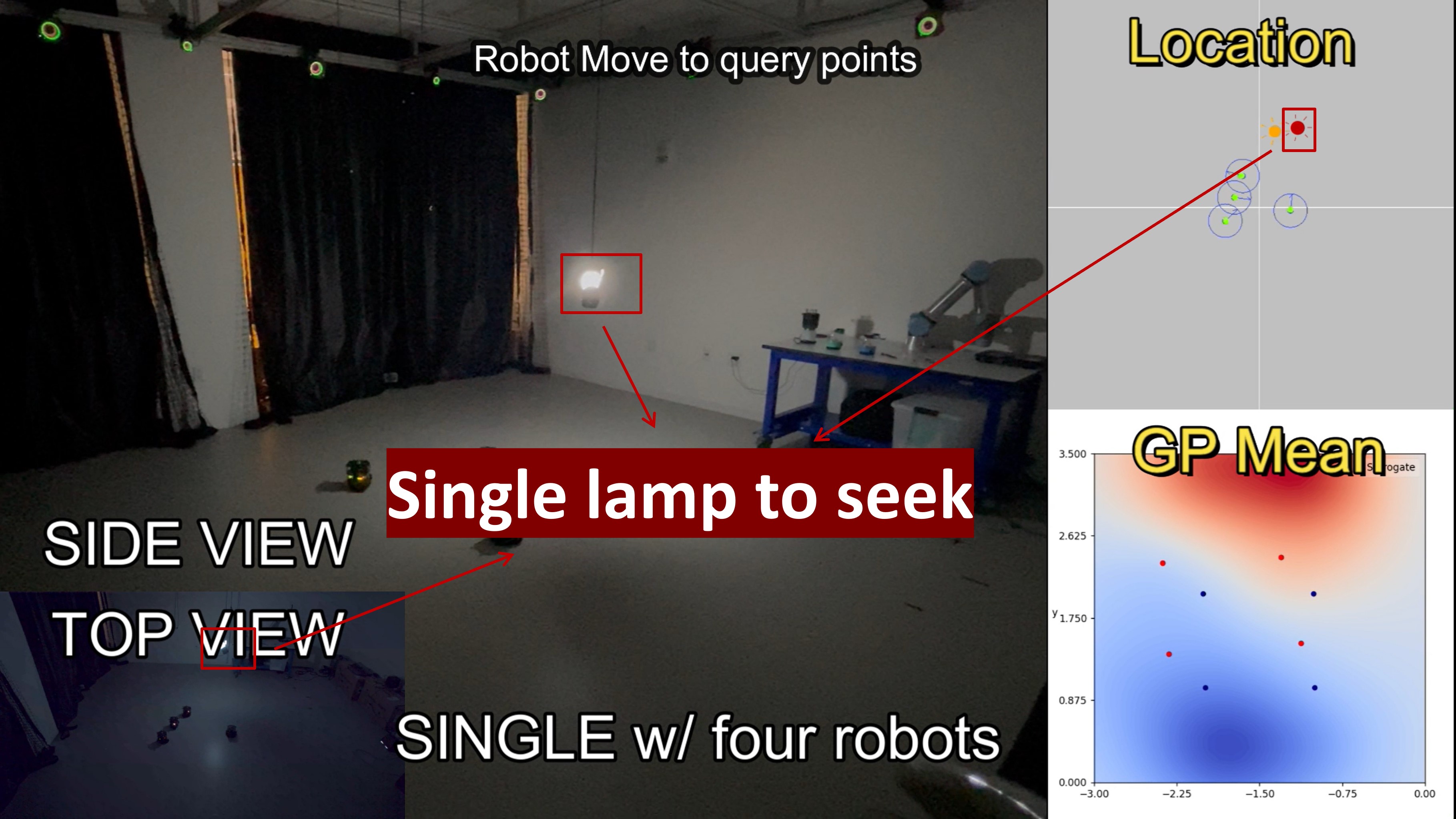}\label{subfig:SINGLE}}
    \subfigure[SPARSE]{\includegraphics[height=0.25\linewidth]{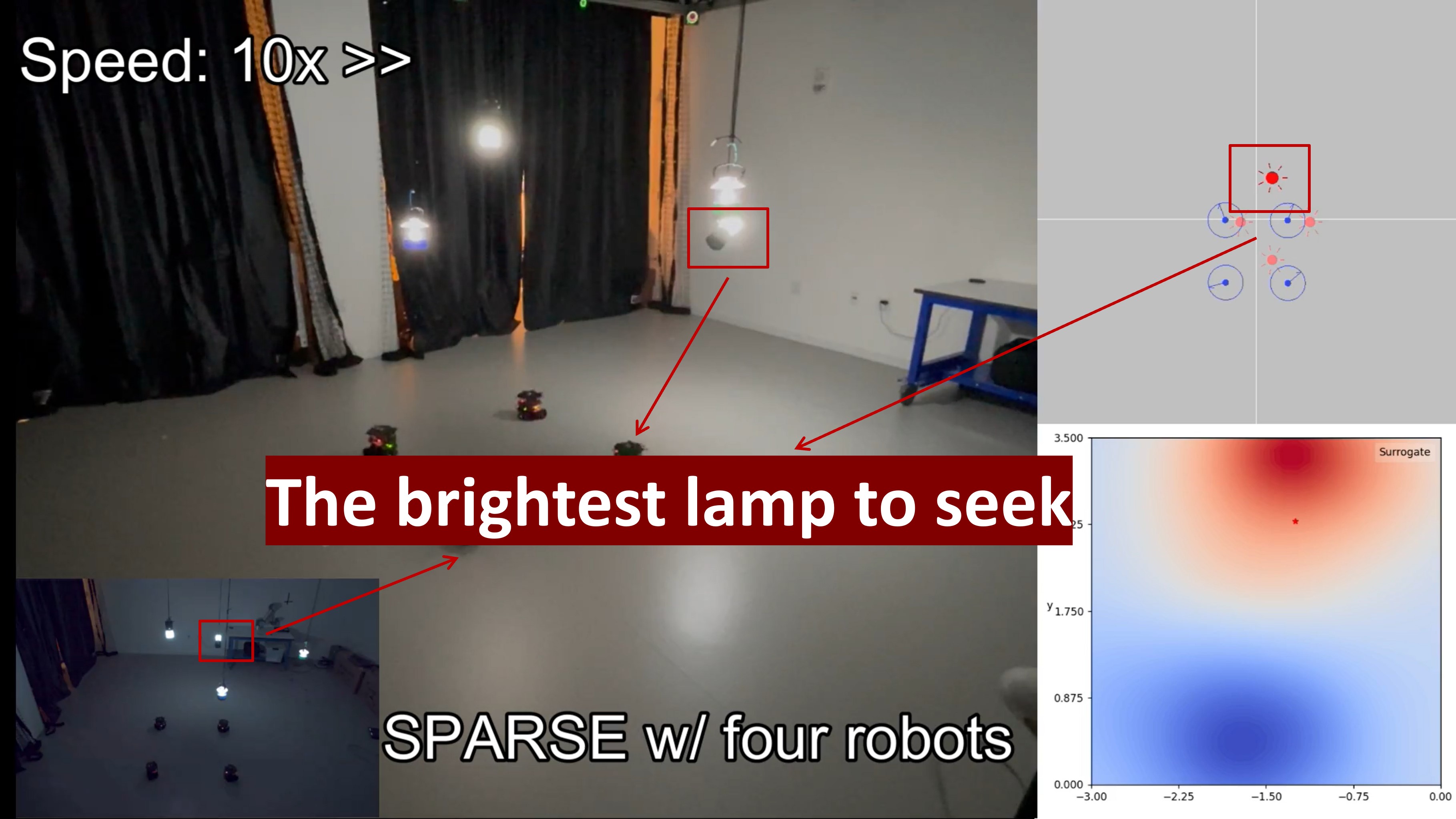}\label{subfig:SPARSE}}
    \subfigure[DENSE]{\includegraphics[height=0.25\linewidth]{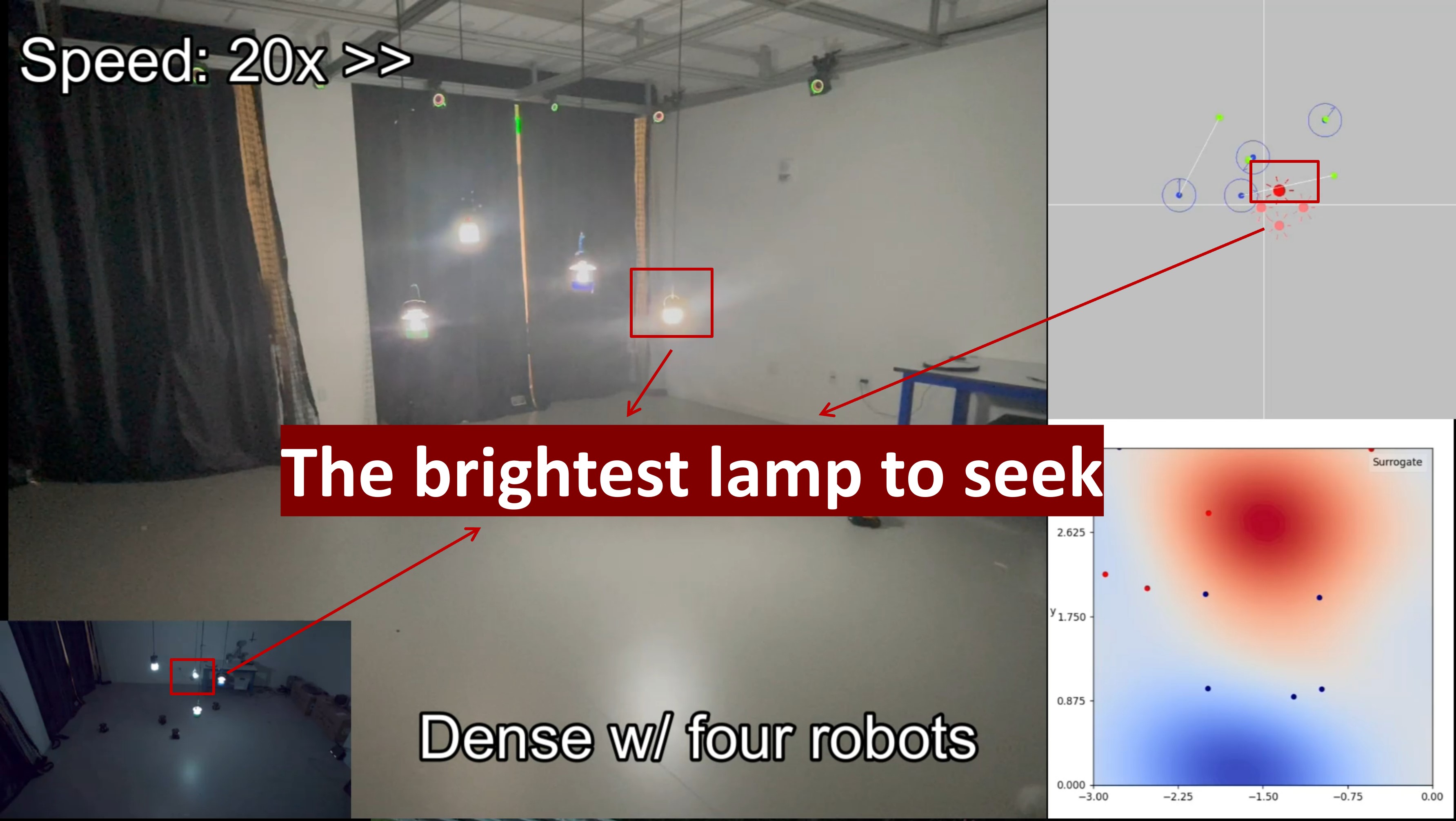}\label{subfig:DENSE}}
     \subfigure[Turtlebot]{
     \includegraphics[height=0.25\linewidth]{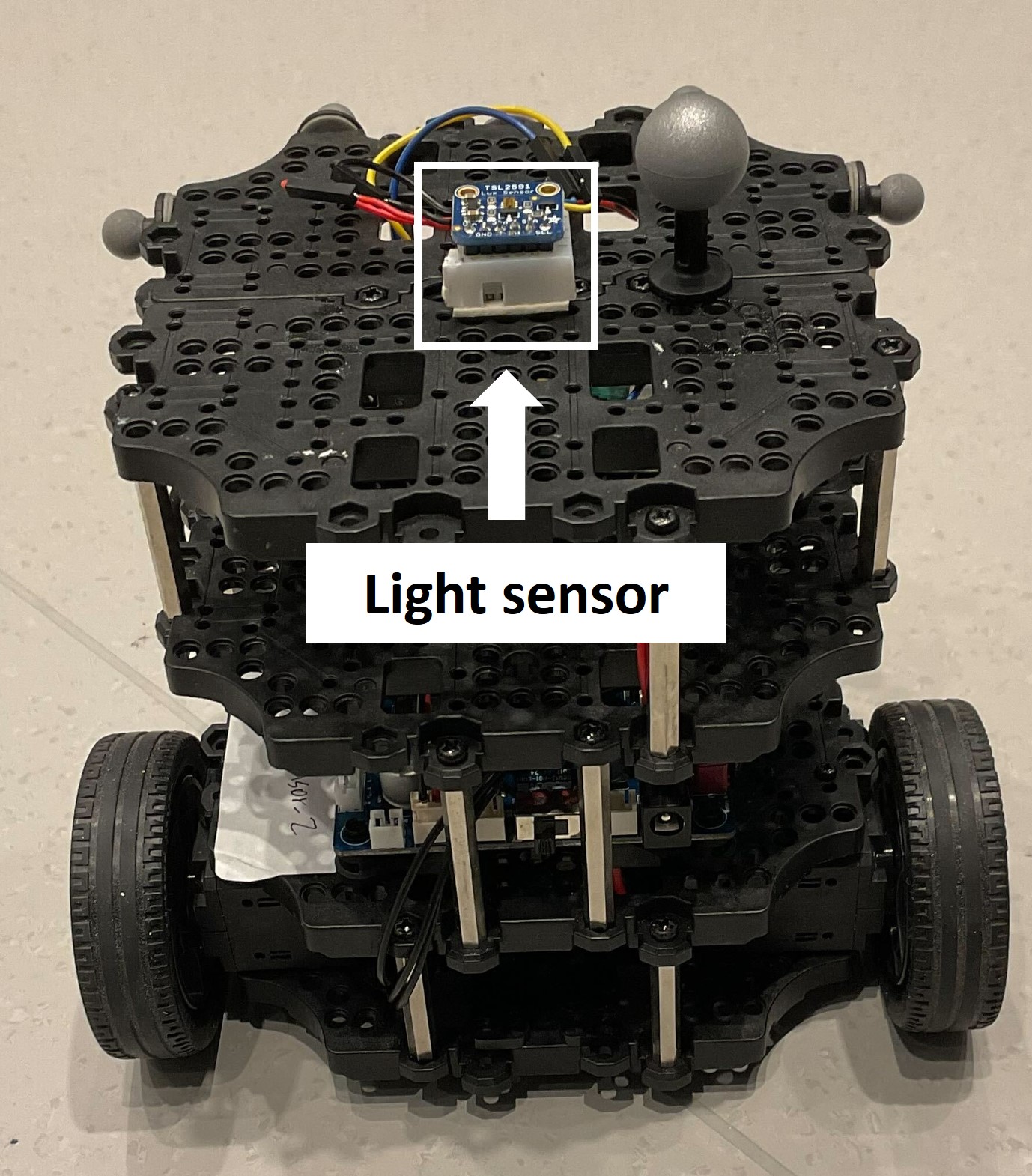}   \label{fig:turtlebot}}
    \caption{Source-seeking experiment setups for different tasks. In each of Figures \ref{subfig:SINGLE}-\ref{subfig:DENSE}: the largest subfigure provides the side and top views of the experiment, where the target lamp to seek is marked by a red box; the subfigure on the upper right illustrates the relative location of the robots (blue circles) and the lights (red, sun-shaped icons) projected to the ground plane; the subfigure on the lower right shows the contour plot about the GP posterior mean at the time when the pictures are taken, where red and blue colors indicate high and low posterior mean values, respectively. Figure \ref{fig:turtlebot} illustrates the light sensor setup on the robot.}
    \label{fig:videoclips}
\end{figure}

In the experiments, we use four ROBOTIS TurtleBot3 with onboard light sensors, as illustrated in Figure \ref{fig:turtlebot}. The goal is to find the location of the highest brightness on the ground level in a dark room. We use a desktop computer as the central coordinator to maintain the GP model and calculate the queries. Figure \ref{fig:videoclips} shows three different source seeking experimental setups. The simplest task, SINGLE, has only one LED lamp hung above the ground, corresponding to the brightest location in the room. The other two tasks, labeled SPARSE and DENSE, are with four lamps in the room, where two of the lamps are brighter and the other two dimmer. Each lamp is hung at a different height, and the brighter lamp hanging closest to the ground, marked by red boxes in Figures \ref{subfig:SPARSE} and \ref{subfig:DENSE}, corresponds to the brightest location in the room. The only differences between the two four-light experiments are the lamp spacing.


We use the basic look-ahead PID controller to drive the Turtlebots to the query points our multi-agent BO algorithm decides while avoiding collision between the robots. See Appendix C of our online report\cite{online_report} for details about our robot controller. We include the log-barrier safety term in the acquisition function to ensure the robots' target positions do not induce collisions. The experiments terminate when the inferred maximum (the location of the maximal posterior mean function $\mu_t$) is within 0.1m of the actual brightest location for three consecutive iterations.
\subsection{Results}

\begin{figure}
    \centering
    \includegraphics[width=\linewidth]{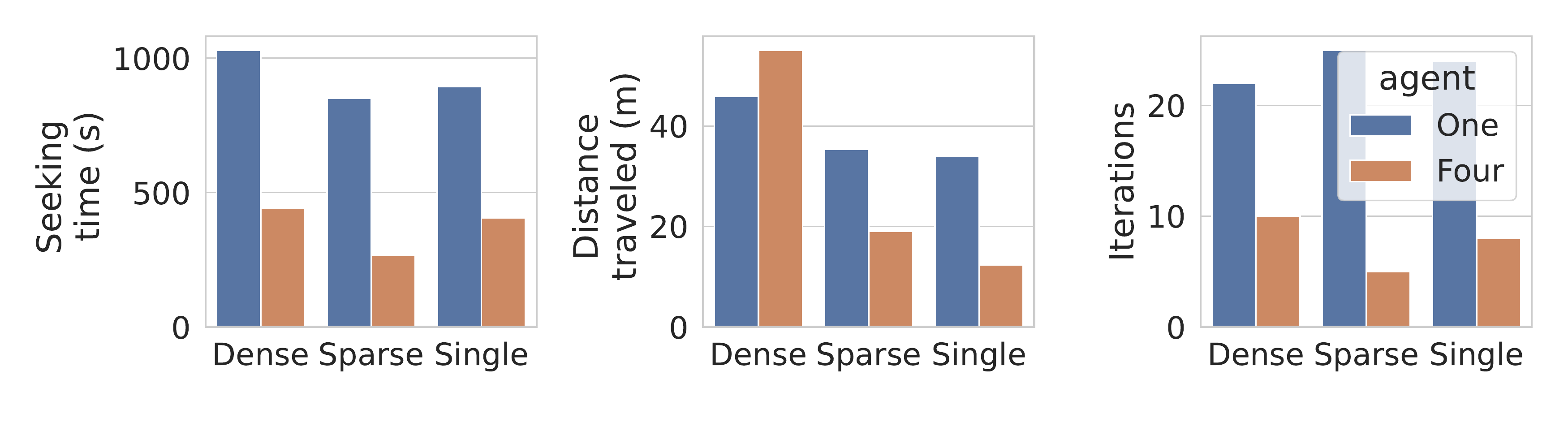}
    \caption{Comparisons between the single- and multi-agent approaches in different environment setups, as described in Figure \ref{fig:videoclips}.}
    \label{fig:real1}
\end{figure}

The accompanying video shows that the robots consistently find the highest brightness location within a short time. The results suggest our algorithm can be applied to a multi-agent team and efficiently maximize a general black-box function in the real world.
Figure \ref{fig:real1} further shows the performance difference of source seeking with four robots compared to using only one robot.
The multi-agent team saves the source seeking time by 59.9\% and the iterations by 67.6\% compared with the single agent. This impressive advantage in efficiency over the single-agent approach shows the multi-agent BO approach can significantly benefit time-critical source seeking applications, such as search-and-rescue missions. More experimental results could be found in Appendix D in the online report \cite{online_report}.

\section{Conclusion}
In this paper, we proposed the Gaussian max-value Entropy Search (GMES), a computationally efficient algorithm for multi-agent Bayesian optimization. We use the normal distribution to approximate the posterior function max-value to design an acquisition function that balances the exploration-exploitation trade-off and has a closed-form expression that allows simple computation without complex approximations and relaxations. We further improve the algorithm by using gradient ascent for the query point optimization and a log-barrier term to enforce safety constraints. Experiment results show that the GMES outperforms other multi-agent BO baselines in the numerical experiments and effectively seeks light sources on real robots. Future works include analyzing the performance of GMES theoretically and using GMES with the distributed multi-agent settings with local communications only.

\bibliographystyle{bib/IEEEtran}
\bibliography{bib/irosref}
\appendices
\section{Proof of Proposition \ref{prop:gamma}}\label{append:prop-gamma-proof}
\begin{proof}

Denote $I_{k}$ as the $k$-dimensional identity matrix for any positive integer $k$.
\begin{equation}
	\begin{aligned}
		&\hat \sigma^2_{t+1}(\xucb)\\
		=& k(\x, \x) - \\
            &\left[\boldsymbol{k}_t(\x)^\top, k(\x,\batchx_t)^\top\right]\times\\
            &\left(\begin{array}{cc}
				(\boldsymbol{K}_t + \sigma_0^2I_{mt} & \boldsymbol{k}_t(\batchx_t) \\
				\boldsymbol{k}_t(\batchx_t)^\top & (k(\batchx_t, \batchx_t) + \sigma_0^2I_m)
			\end{array}\right)^{-1}\times\\
            &\left(\begin{array}{c}
				\boldsymbol{k}_t(\x) \\
				k(\batchx_t, \x)
			\end{array}\right)  \\
		& \text{(using matrix inverse lemma)}\\
		=& k(\x, \x) - \boldsymbol{k}_t(\x)^\top (\boldsymbol{K}_t + \sigma_0^2I_{mt})^{-1} \boldsymbol{k}_t(\x) \\
		&-\left(k(\x,\batchx_t)-\boldsymbol{k}_t(\x)^\top (\boldsymbol{K}_t + \sigma_0^2I_{mt})^{-1} \boldsymbol{k}_t(\batchx_t)\right) \times \\
		&\left((k(\batchx_t, \batchx_t) + \sigma_0^2I_m)-\boldsymbol{k}_t(\batchx_t)^\top (\boldsymbol{K}_t + \sigma_0^2I_{mt})^{-1} \boldsymbol{k}_t(\batchx_t)\right)^{-1}\times\\
		&\left(k(\batchx_t, \x)-\boldsymbol{k}_t(\batchx_t) ^\top(\boldsymbol{K}_t + \sigma_0^2I_{mt})^{-1} \boldsymbol{k}_t(\x)\right) \\
		=& \sigma^2_t(\x)-\Sigma_t\left(\x, \batchx_t\right) (\Sigma_t^{-1}(\batchx_t, \batchx_t)+\sigma_0^2I_m)\Sigma_t\left(\batchx_t, \x\right)
	\end{aligned}
\end{equation}	
\hfill$\blacksquare$
\end{proof}


\section{Multi-agent Gradient Ascent Gaussian Max-value Entropy Search}\label{append:alg-gd}

\begin{algorithm}[htp]  
    \caption{Multi-Agent Gradient Ascent Gaussian Max-value Entropy Search
    }\label{alg:gd}
    
    \LinesNumbered
    \Parameter{
    Gradient ascent step $N$,
    Step size sequence $\delta_t$,
   other parameters initialized the same way as in Algorithm \ref{alg:ori}.
    }
    \KwOut{Inferred function maximum from $GP(\mu_T,\Sigma_T\mid D_T)$}
    \For{$t = 1,2,\dots,T$}
    {
    \algcomments{Agent observe objective functions.}
    
    The agents query $\batchx_{t-1}$ and observe $\batchy_{t-1}$. 
    
  \algcomments{The central coordinator computes the new query points $X_t$ through gradient ascent.}     
    
    \algcomments{The central coordinator computes the new query points $X_t$ through gradient ascent.} 
 $\xucb_t\leftarrow\argmax_{\x \in \mathcal{X}} \mu_t(\x) + \beta_t\sigma_t(\x)$

    Initialize $\batchx_t$ so that $\|\x_t^i-\x_t^j\| \geq r_{\text{div}}$ for all $i\neq j$. 
    
    \For{$n = 1,2,\dots, N$}
        {
                
        $\batchx_t\leftarrow \Gamma_{\mathcal{X}^m}\big(\batchx_t + \delta_t\cdot\nabla_{\batchx_t}\left(\gamma(\batchx_t, \xucb_t) - p(\batchx_t)\right)\big)$ 
        }

       Central coordinator publishes new query points $X_t$ to the agents. 
    }
\end{algorithm}


\section{Low-level Robot Controller and Collision Avoidance in Source Seeking Experiments}\label{append:low-level-control}
\begin{figure}[ht]
    \centering
    \includegraphics[width=0.8\linewidth]{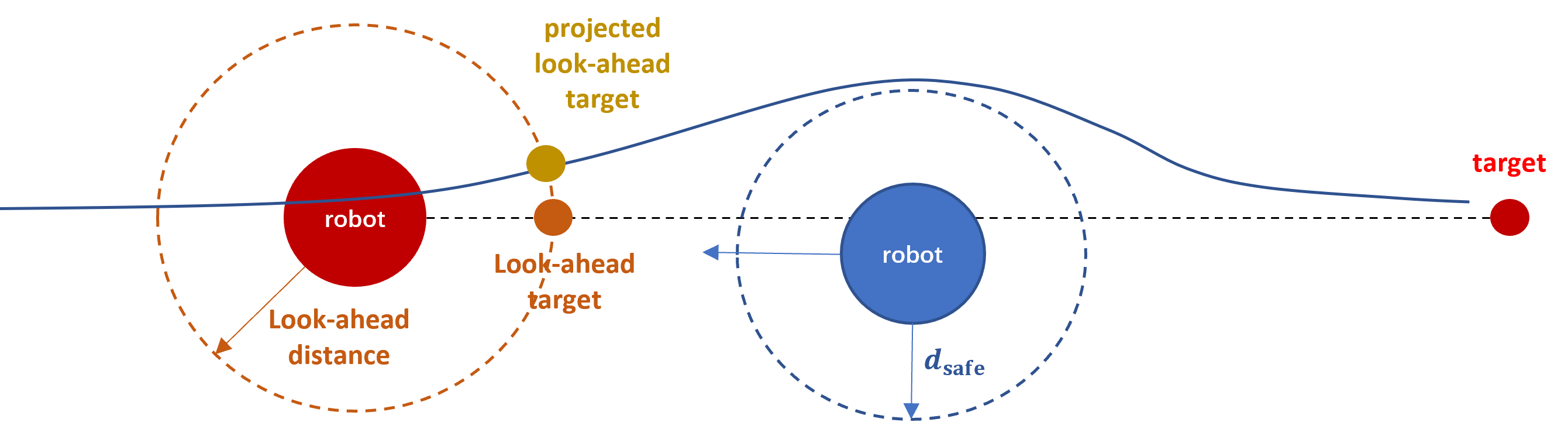}
    \caption{Look-ahead target projection for collision avoidance. }
    \label{fig:target_proj}
\end{figure}

We use the basic look-ahead PID controller to drive the Turtlebots to target positions, which are the query points decided by our multi-agent BO algorithm. As shown in Figure \ref{fig:target_proj}, the target position given by the multi-agent BO algorithm is projected to the \textit{look-ahead target position}, whose distance to the robot is limited by the look-ahead distance $r_{\text{lk}}$. The look-ahead distance is linearly related to the linear velocity of the robot, defined by 
\begin{equation}
    r_{\text{lk}} = v t_{\text{lk}} \label{eq:lka}
\end{equation} 
where $v$ is the current linear velocity and $t_{\text{lk}}$ is the lookahead time.
Then the look-ahead target is computed by solving the convex optimization problem \eqref{eq:target_proj}

We also don't want the robots to collide into each other. We use the target point projection methods based on control barrier function (CBF) \cite{ames2019control,ma2021model} to avoid collision between each robots. For agent indexed by $i$ and other agents indexed by $j$, the target projection is formulated by the following convex optimization problem,
\begin{equation}
\begin{aligned}
    \min_{\x_i\in\mathcal{X}} & \|\x_i -  \x^{\text{lookahead}}_i\|\\
    &\dot d(\x_i,\x_j)\geq -\alpha(d(\x_i,\x_j) - d_{\text{safe}})\  \text{for all }j \neq i
\end{aligned} \label{eq:target_proj}
\end{equation}
where $\mathbf{p}_j$ is the current position of robot $j$, and $d(\x_i,\x_j) = \|\x_i-\x_j\|_2$ is the distance between the look-ahead target of agent $i$ and current position of , $\dot d(\x_i,\x_j)$ is the derivative with respect to time, $d_{\text{safe}}$ is a safe distance, and $\alpha(\cdot)$ is a class $\mathcal{K}$ function, and we simply use a linear function here by $\alpha(x) = k_\alpha x$. All the hyper-parameters are listed in Table \ref{tab:para}.

\begin{table*}[ht]
\centering
\caption{Hyper-parameters in the Source Seeking Experiment} \label{tab:para}
\begin{tabular}{ccc}
\hline
Notation        & Meaning                                   & Value (unit if there is) \\ \hline
$d_\text{safe}$ & Safe distance used in \eqref{eq:target_proj}            & 0.2                      \\
$k_\alpha$      & coefficent in the class $\mathcal{K}$ function & 0.1                      \\
$t_{\text{lk}}$ & Lookahead time in \eqref{eq:lka}          & 1.0 (s)   \\
\hline
\end{tabular}
\end{table*}

\section{Additional Experimental Results}
\subsection{Numerical Simulations}
\subsubsection{Experiment Results of Single-Agent BO}\label{append:single-agent-BO}

\begin{figure}[ht]
	        \centering
	        \setcounter{subfigure}{0}
	        \subfigure{\includegraphics[width=0.45\linewidth]{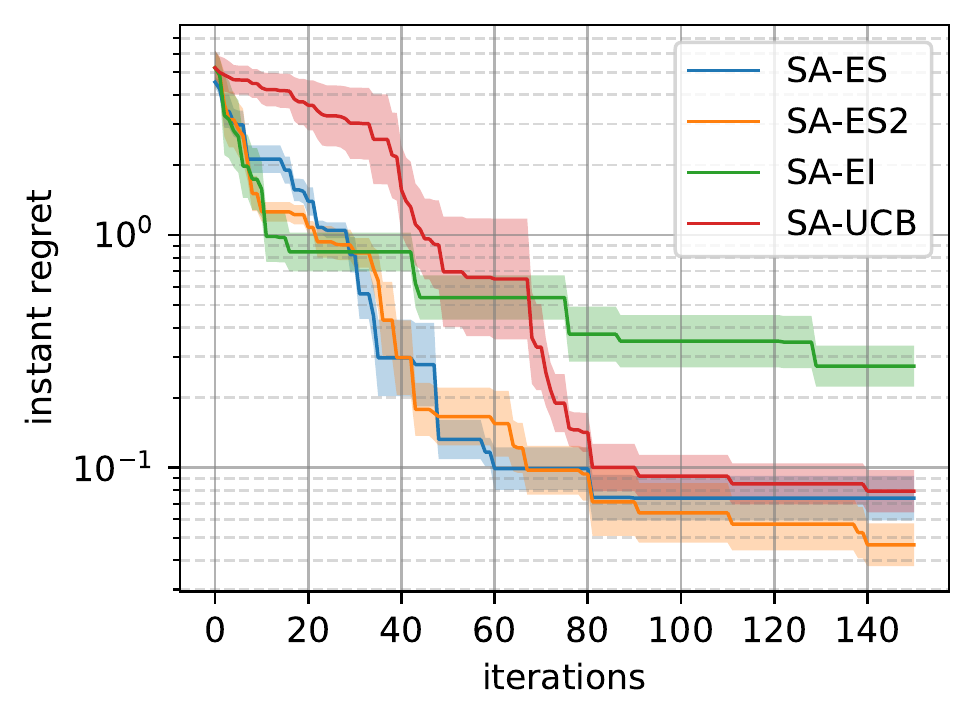}}
	        \subfigure{\includegraphics[width=0.45\linewidth]{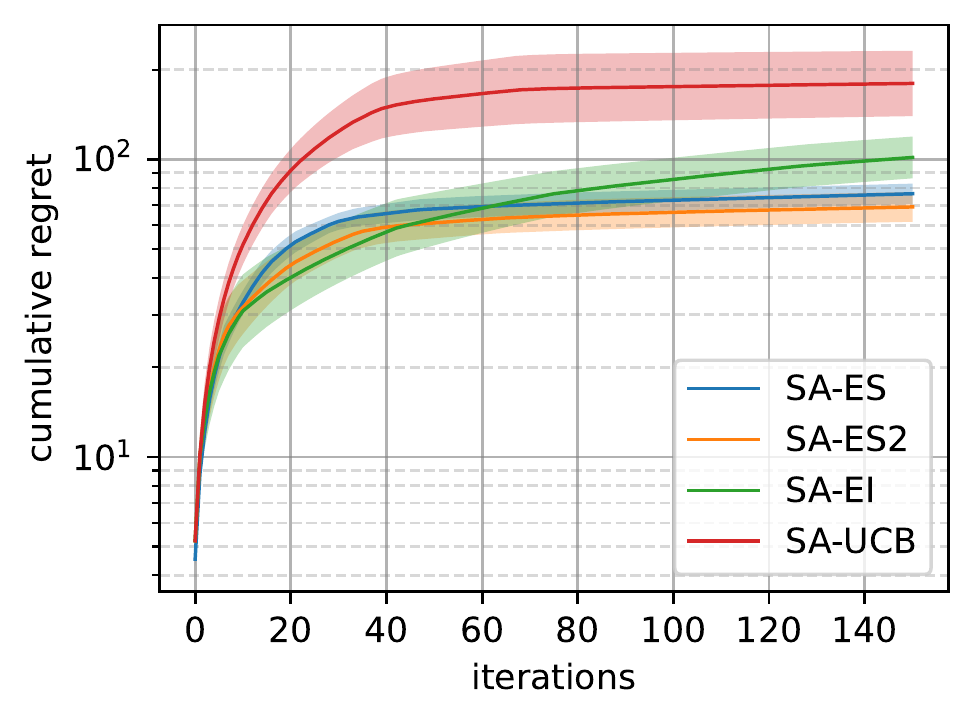}}
         \caption{Regret of Ackley task for single-agent tasks.}
         \label{fig:single}
	    \end{figure}

The performance of proposed algorithms on the single-agent BO problem is presented in Figure~\ref{fig:single}. We only report the results on the Ackley task, though the results on other test functions are similar. In the single-agent problem, we can afford to discretize the search space $\mathcal{X}$ and carry out the maximization of $\gamma$ in our algorithms through brute force. In Figure \ref{fig:single}, ES denotes the performance of our algorithm when $\batchx_t = \argmax_{\batchx \in \mathcal{X}} \gamma(\batchx,\xucb_t)$ is computed with brute force, while ES2 the performance when $X_t$ is computed through gradient ascent. We also consider two baseline algorithms: expected improvement (SA-EI) and upper confidence bound (SA-UCB). We observe that ES and ES2 perform very similarly under both metrics and have noticeably better empirical performance than the baselines. These results suggest that using gradient ascent to determine $X_t$ as in Algorithm \ref{alg:gd} can still achieve comparable BO performance as computing $X_t$ through brute force. 
 
\begin{figure}[htp]
    \centering
    \subfigure[Ackley]{\includegraphics[width=0.48\linewidth]{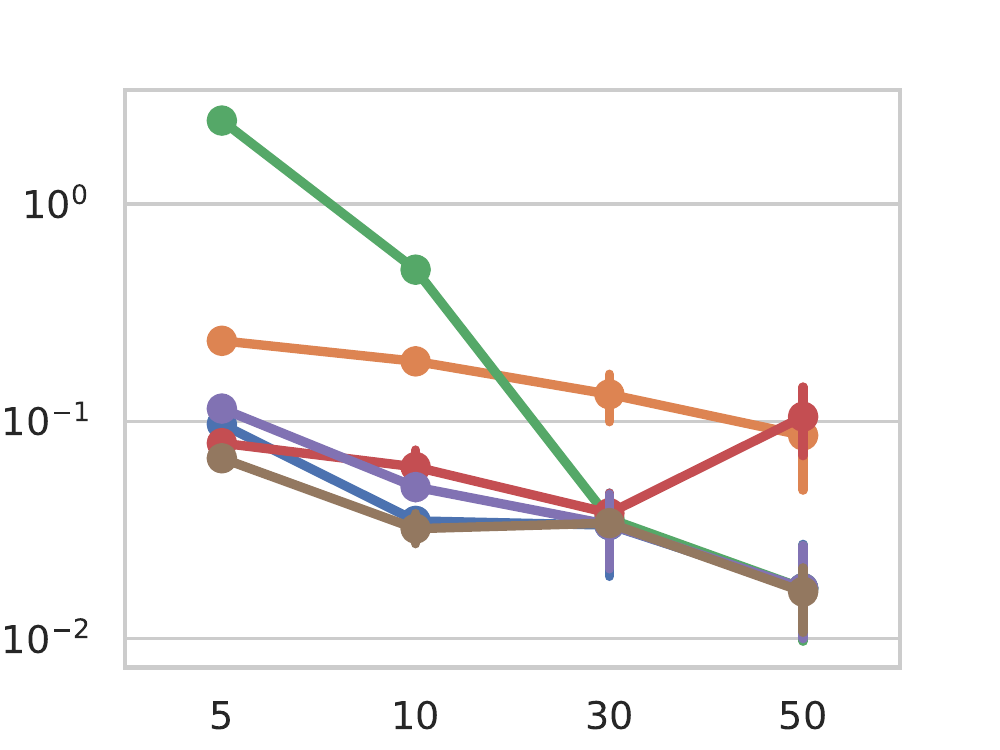}}
    \subfigure[Bird]{\includegraphics[width=0.48\linewidth]{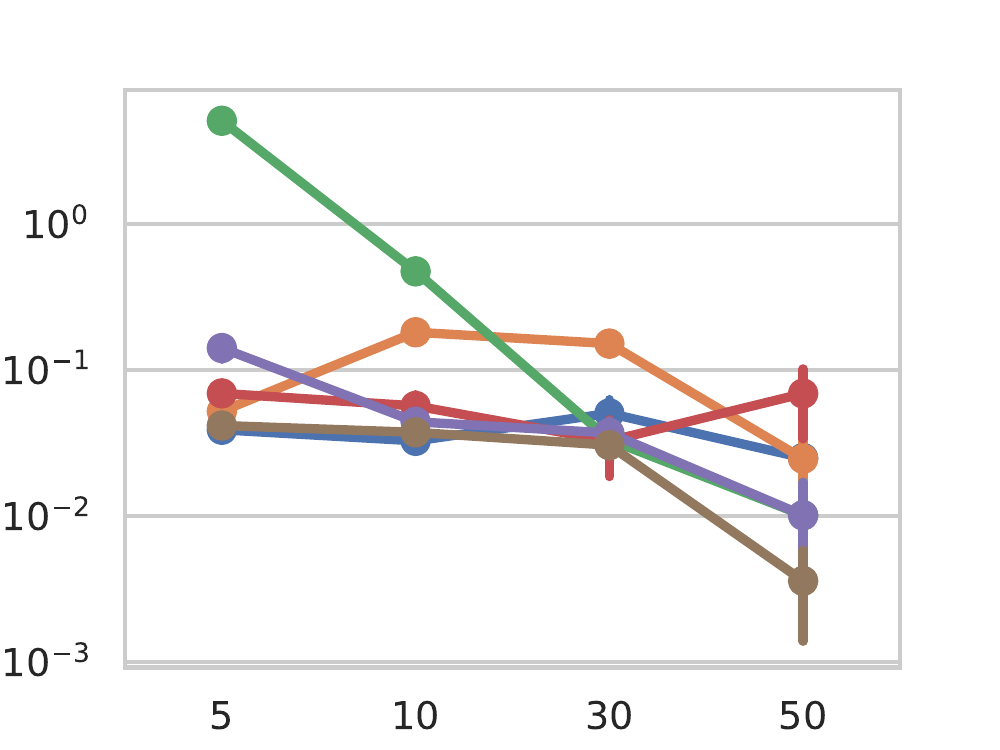}}
    \subfigure[Rosenbrock]{\includegraphics[width=0.48\linewidth]{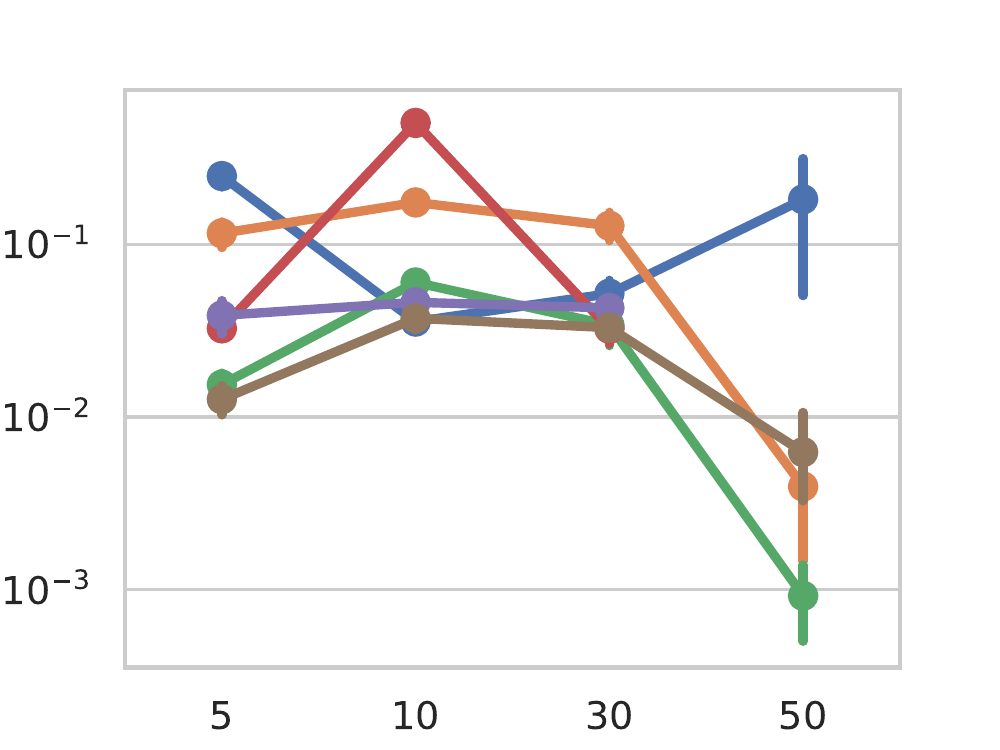}}
    \includegraphics[width=0.28\linewidth]{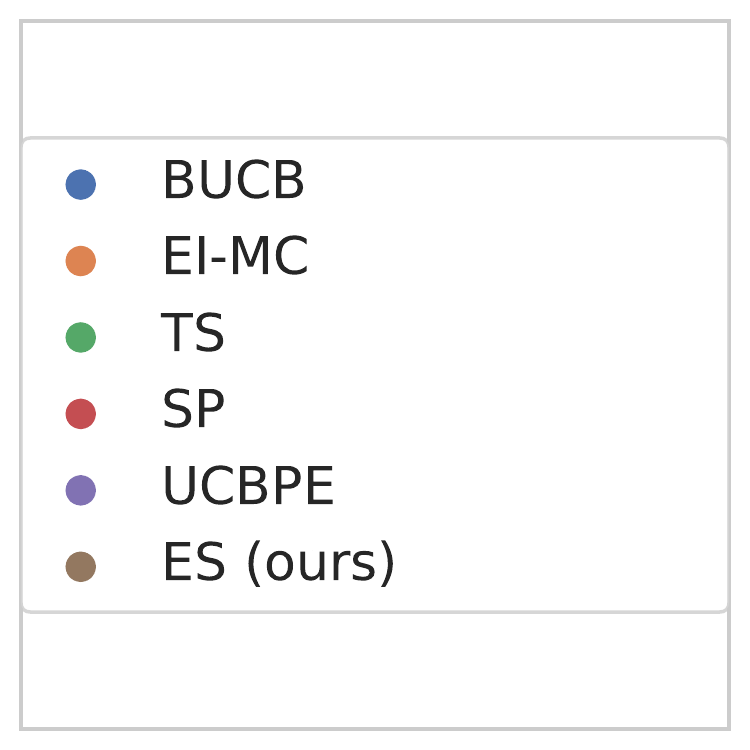}
    \caption{Instant regret at the last step (150 step in total) with different agents. Dots and lines are the average and 95 percent confidence interval among 5 runs. }
    \label{fig:regret_agents}
\end{figure}
\subsubsection{Algorithm Performance Change with Agents}
We also implement our algorithm with 10, 30, 50 agents to see how our algorithm performs with larger numbers of agents, and we plot the results in Figure \ref{fig:regret_agents}. The results show that for most of cases with different objective functions and agents, ES performs the best compared to all the baselines. For some cases like Bird function with 10 agents or Rosenbrock function with 50 agents, the performance is not the best but still the second or third best. Meanwhile, we can see that on Ackley and Bird, the performance of ES increase with the number of agents, which shows that the collaboration effect of multiple agents. On Rosenbrock, the effect is not that obvious but the performance with 50 agents are still much better than 5, 10, and 30 agents.

\subsection{Source Seeking Physical Experiments}
\begin{figure}[htb]
    \centering
    \includegraphics[width=\linewidth]{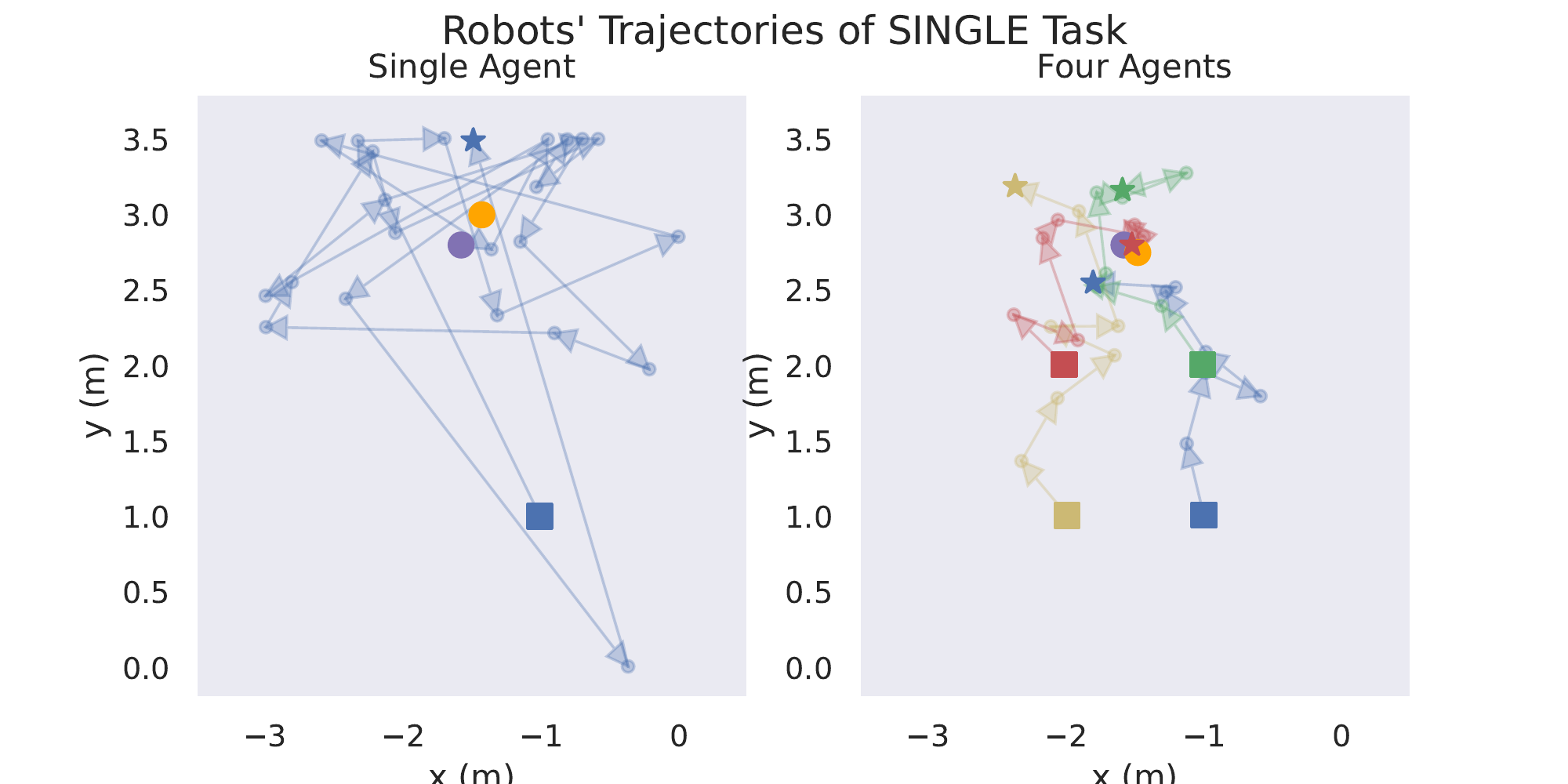}
    \includegraphics[width=\linewidth]{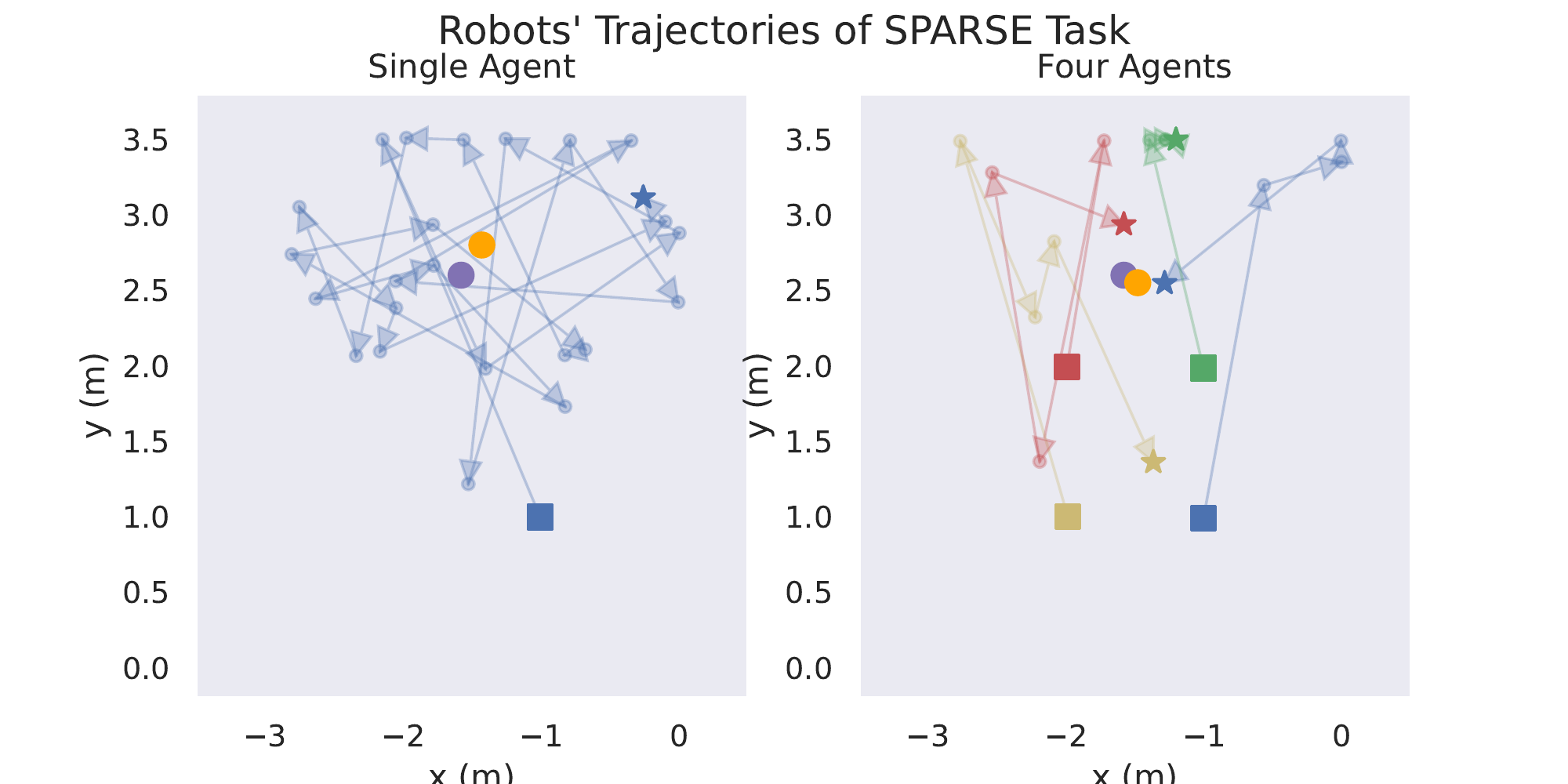}
    \includegraphics[width=\linewidth]{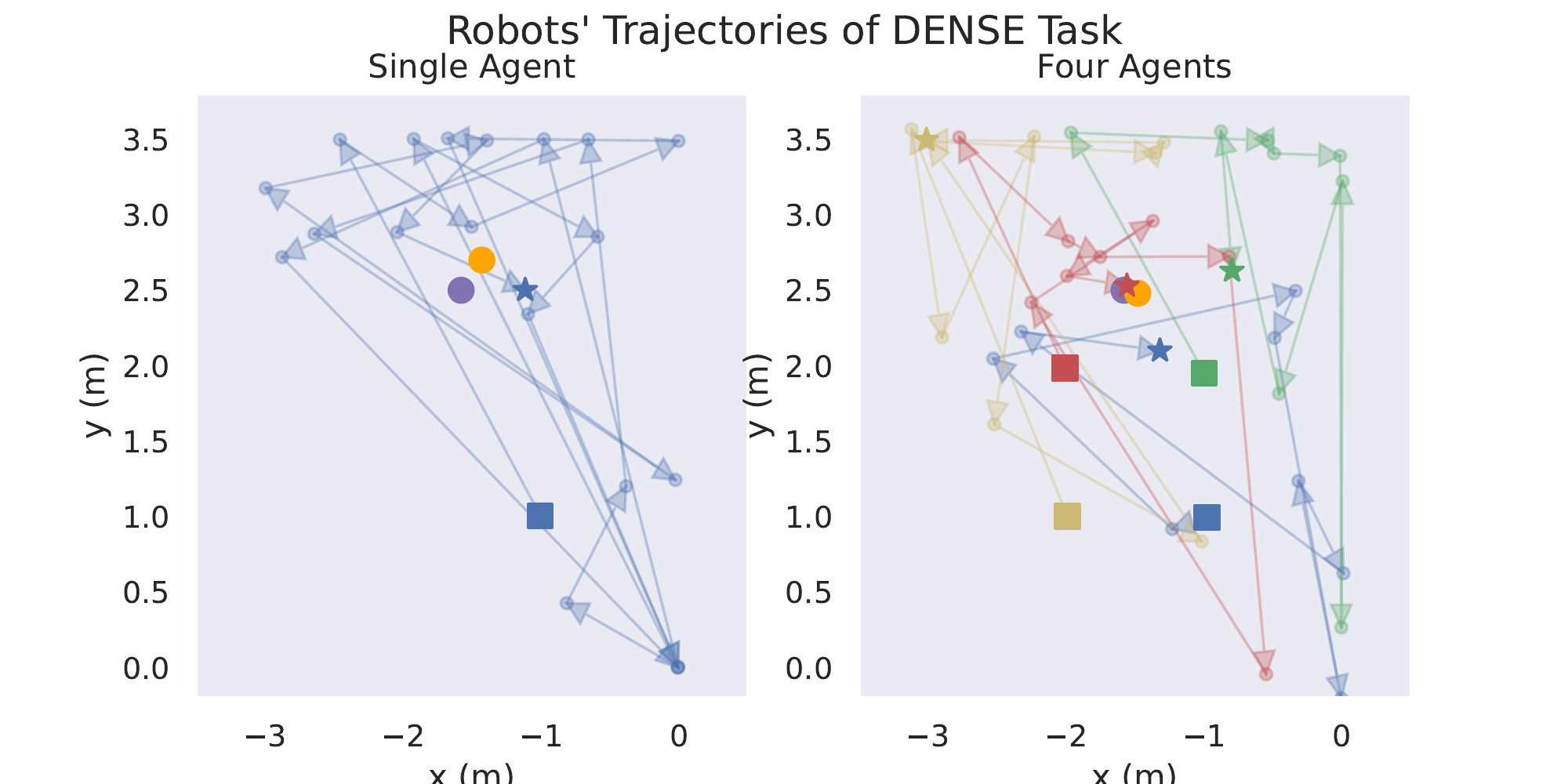}
    \includegraphics[width=\linewidth]{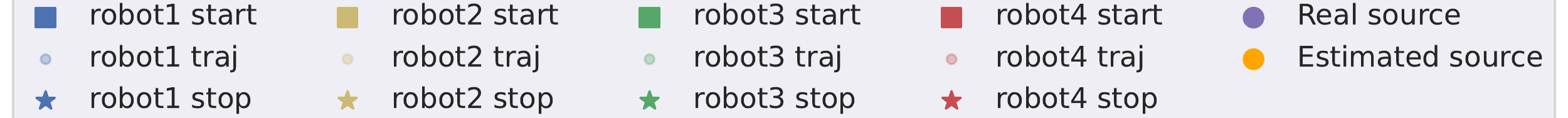}
    \caption{Robot trajectories and queries of the source seeking problem.}
    \label{fig:sstraj}
\end{figure}

\end{document}